\documentclass[amsmath,amstext,amssymb,amsfonts,10pt,a4paper,twocolumn,nobalancelastpage,aps,physrev,longbibliography]{revtex4-2}

\usepackage[english]{babel}
\usepackage{color,graphicx}
\usepackage{upgreek}
\usepackage{dcolumn}
\usepackage{bm}
\usepackage[per-mode=symbol]{siunitx}
\usepackage{array}

\usepackage[colorlinks=true, linkcolor=blue]{hyperref} 

\usepackage{cleveref}
\usepackage{color}
\usepackage{calrsfs}

\newcommand{\op}[1]{{\mathcal{#1}}}
\newcommand{\dd}[1]{\mathrm{d}\,#1}

\newcommand{\opt}{\mathrm{opt}}

\newcommand{\exclude}[1]{-{#1}}
\newcommand{\avg}[1]{\left\langle #1 \right\rangle}
\newcommand{\nn}{\mathrm{nn}}
\newcommand{\train}{\mathrm{train}}
\newcommand{\alternate}[1]{\widetilde{#1}}
\newcolumntype{L}{>{$}l<{$}}
\newcolumntype{R}{>{$}r<{$}}
\newcolumntype{C}{>{$}c<{$}}

\begin{document}

\title{\textit{Nash Neural Networks}: Inferring Utilities from Optimal Behaviour}
\bigskip
\author{John J. Molina}
\email[]{john@cheme.kyoto-u-ac.jp}
\author{Simon K. Schnyder}
\affiliation{Department of Chemical Engineering, Kyoto University, Kyoto 615-8510, Japan}
\author{Matthew S. Turner}
\affiliation{Department of Physics, Warwick University}
\affiliation{Department of Chemical Engineering, Kyoto University, Kyoto 615-8510, Japan}
\author{Ryoichi Yamamoto}
\affiliation{Department of Chemical Engineering, Kyoto University, Kyoto 615-8510, Japan}

\date{\today}

\begin{abstract}
	We propose Nash Neural Networks ($N^3$) as a new type of Physics Informed Neural Network that is able to infer the underlying utility from observations of how rational individuals behave in a differential game with a Nash equilibrium. We assume that the dynamics for both the population and the individual are known, but not the payoff function, which specifies the cost per unit time of being in any particular state. We construct our network in such a way that the Euler-Lagrange equations of the corresponding optimal control problem are satisfied and the optimal control is self-consistently determined. In this way, we are able to learn the unknown payoff function in an unsupervised manner. We have applied the $N^3$ to study the optimal behaviour during epidemics, in which individuals can choose to socially distance depending on the state of the pandemic and the cost of being infected. Training our network against synthetic data for a simple SIR model, we showed that it is possible to accurately reproduce the hidden payoff function, in such a way that the game dynamics are respected. Our approach will have far-reaching applications, as it allows one to infer utilities from behavioural data, and can thus be applied to study a wide array of problems in science, engineering, economics and government planning.
\end{abstract}
\maketitle

\section{Introduction}\label{s:intro}
Differential games are used to analyze situations in which individual players seek to maximize (minimize) their own utilities (losses), in the presence of other players. These ``games'' are not only central to our understanding of social, economic and political planning and processes\cite{Aumann1992,Myerson1997,Dockner2000,Lambertini2018}, but also crucial in biology\cite{McNamara2020}, engineering\cite{Bauso2016}, and computer science\cite{Nisan2007}. Among the most useful concepts in game theory, is that of a Nash equilibrium, which refers to the situation from which no individual can gain any advantage by unilaterally modifying their strategy\cite{Bauso2016}. In other words, if $U_i$ is the utility of player $i$ ($1\le i \le N$), with control/strategy variable $k_i$, then, the Nash equilibrium solution $\bm{k}^\ast = (k_1^\ast,\cdots, k_N^\ast)$, is such that\cite{Bauso2016}
\begin{align}
	U_i(k_i^\ast; \bm{k}_{\exclude{i}}^\ast) & \ge U_i(k_i; \bm{k}_{\exclude{i}}^\ast)\qquad \forall i
\end{align}
where $\exclude{i}$ indicates that the $i$-th coordinate is missing. This state need not be equivalent to the global maximum of the population's utility $U=\sum_i U_i$, i.e., the utilitarian maximum. In this work, we develop neural networks capable of solving inverse problems in the context of differential games with Nash equilbria (i.e., game theoretic inverse optimal control). More specifically, given observations of the behavior of rational individuals within a differential game, we wish to infer the underlying utility $U_i$ from which their decision making process is derived.

Recent advances in machine learning for the physical sciences\cite{Brunton2019} have seen an explosion in the development of Physics Informed Neural Networks (PINN), in which the laws of physics are incorporated into the learning as inductive biases\cite{Raissi2019a,Karniadakis2021}. The results are typically orders of magnitude better than those of naive or baseline networks, in which no such additional structure is included. Not only do PINNs provide better predictions, but they do so while requiring less training data, and generally satisfying the conservations and symmetries of the system under study. The basic idea of such Neural Networks (NN) is quite simple: Instead of directly trying to learn the output $y=f(x)$ from the input $x$, one looks to exploit the known physical laws. These are typically in the form of a differential equation, and either (1) express $f(x)$ in terms of a more fundamental function, thus encoding the physics into the structure of the network, (e.g., Euler-Lagrange equations in terms of the Lagrangian) and/or (2) define the loss function such that all known equations are (approximately) satisfied. Recent examples, which have motivated our current study, include the family of Hamiltonian and Lagrangian Neural Networks\cite{Greydanus2019, Bertalan2019,Zhong2019,Cranmer2020, Zhong2020, Choudhary2020, Lee2020,Finzi2020,Roehrl2020,Lutter2021, Zhong2021,Duong2021,Han2021,Sosanya2022,Celledoni2022,Chen2022}, in which Hamiltonian/Lagrangian mechanics is directly encoded into the neural network. This is achieved by defining a base neural-network to approximate the Hamiltonian (Lagrangian) and deriving from it the corresponding neural-networks that predict the dynamical equations of motion. The training can then be performed directly on the observed dynamical behaviour, one never needs to measure the unknown Hamiltonian or Lagrangian, it is learned in an unsupervised manner. Other relevant examples include Symplectic NN\cite{Jin2020, Chen2019, Meng2022}, Lipschitz Recurrent NN\cite{Erichson2020}, Poisson NN\cite{Jin2022}, GENERIC formalism informed NN\cite{Sipka2021,Zhang2021a}, and Noether NN\cite{Alet2021}, among others.

Deep learning for optimal control problems has typically focused on overcoming the ``curse of dimensionality'' that is encountered when solving problems with many agents (high-dimensions).
Several strategies have been developed to overcome this issue: directly learning the solution (trained to satisfy the known constraints)\cite{Sirignano2018}, learning the value function (i.e., the integrated payoff function evaluated at the optimal control)\cite{Niarchos2006,Djeridane2006,Jiang2017,Royo2017,Nakamura-zimmerer2020,Nakamura-zimmerer2020a,Onken2021,Bansal2021}, recasting the corresponding Hamilton-Jacobi partial differential equations as backward stochastic differential equations to learn the gradient of the solution\cite{Han2018,Hure2020,Bachouch2022}, or leveraging neural ordinary differential equations to automatically learn control signals\cite{Bottcher2022}.

The inverse problem, that of learning the underlying cost or payoff functions has also been extensively studied. This was usually done by assuming some fixed form for the function, in terms of basis functions or feature vectors, and then learning the corresponding weights or function parameters\cite{Mombaur2010, Puydupin-Jamin2012, Englert2017, Molloy2018,Jin2021b,Arora2021,Cao2022}. A recent approach, closely related to this work, is that of (discrete/continuous) Pontryagin Differentiable Programming (PDP)\cite{Jin2020a,Jin2022}, which introduces the optimal control theory into the learning framework, providing end-to-end differentiable learning and control. PDP has been applied to solve both direct and inverse control problems for robot maneuvering and rigid body motion. When studying the inverse problem, the authors have considered both learning the parameters of a known cost function and learning an unknown function (represented by a neural network). However, for the later, they require that this cost function be separable into an unknown state-dependent term and a known control-dependent term. For robotic manipulation or path planning, this is not an issue, as the control cost is typically known. However, for general inverse optimal control problems this cannot be assumed. In fact, this PDP formalism has been recently applied to a game-theoretic inverse learning problem\cite{Cao2022}, like the one we are considering, but the full form of the cost/payoff function was specified in advance (only the weights/parameters were learned). Finally, connections between specific neural network architectures and the solution to certain Hamilton-Jacobi equations have been found\cite{Darbon2020,Darbon2021,Darbon2021a}. In ref.\cite{Darbon2020}, the authors consider select inverse problems, but their approach requires the Hamitonian to be of a specific form, and they reported that such ``problems cannot generally be solved with the Adam optimizer with high accuracy''.

In this work, we propose Nash Neural Networks ($N^3$) as a physics informed framework to tackle general inverse problems in differential games with Nash equilibria\cite{Isaacs1965,Basar1999,Nash1951}. In particular, we consider the case of a population of (identical) rational individuals that wish to maximize their individual utility. While the form of the dynamical laws governing the dynamics of the population as a whole are assumed fixed, the individuals are able to influence their own time evolution through a control parameter that encodes their behaviour.  The theoretical framework for solving such problems is well established within the variational principles of Classical Mechanics\cite{Arnold1989, Goldstein2001,sicm}, in general, and control theory\cite{Lenhart2007,Lambertini2018}, in particular. We show that the proposed $N^3$ allows us to ``learn'' the underlying utility/cost/payoff function (i.e., the potential energy) defining the individual (population) behaviour. This is done by encoding the network with Lagrangian/Hamiltonian mechanics, self-consistently determining the optimal control (which requires evaluating the network on itself), and assigning this Nash solution to all individuals. In contrast to the PDP approach\cite{Jin2020a,Jin2022,Cao2022}, which does not explicitly encode the optimality condition into the network structure, we make no assumptions regarding the form of the cost function.

This paper is organized as follows: first, we provide a brief overview of the general theory, leaning heavily on the analogy with classical mechanics, as applied to differential games. We then show how this can be used in the context of determining optimal social distancing policies during a pandemic. Finally, we introduce the Nash Neural Networks, and show how they are able to recover the utility of rational individuals within the pandemic example, from observations of their behaviour.

\section{Theoretical Background}\label{s:theory}
In classical mechanics a realizable path $q(t)$, between two fixed times $t_1$ and $t_2$, is such that it extremizes the action functional\cite{Arnold1989,Goldstein2001,sicm}
\begin{align}
	S[q](t_1,t_2) & = \int_{t_1}^{t_2} L(t, q(t), \dot{q}(t), \cdots)\dd{t},\label{e:S}
\end{align}
with $L$ the system Lagrangian, which is generally a function of time, coordinates $q$, and velocities $\dot{q}$, though higher order time-derivatives can also be included. This stationary action principle states that the action is stationary with respect to small variations in the realizable path $q(t) + \epsilon\delta\eta(t)$, i.e., in the limit when $\epsilon\rightarrow 0$, $\delta_q S = 0$, where\cite{Gelfand2000}
\begin{align}
	\delta_q S[q](t_1,t_2) & = \delta_q\int_{t_1}^{t_2} L(t, q(t),\dot{q}(t), \cdots)\,\dd{t} \label{e:dS0}                                                                                                      \\
	                       & = \left.\left(\partial_{\dot{q}} L\right)\delta \eta\right\rvert_{t_1}^{t_2} + \int_{t_1}^{t_2}\left(D_t \partial_{\dot{q}}L - \partial_q L\right)\delta{\eta}\,\dd{t},\label{e:dS}
\end{align}
and $D_t$ is the total-time derivative, defined as\cite{sicm}
\begin{align}
	D_t & \equiv \partial_t + \dot{q}^i\cdot \partial_{q^i} + \ddot{q}^{\,i} \cdot\partial_{\dot{q}^i} + \cdots .
\end{align}
Due to the fact that the variation in the path is arbitrary, except possibly at the end-points, where it should vanish if boundary-conditions are specified for $q(t)$, not only must both terms on the rhs of eq.~\eqref{e:dS} equal zero, but the term in parenthesis in the integrand must also vanish. This gives rise to the well-known Euler-Lagrange equations\cite{Arnold1989,Goldstein2001,sicm}
\begin{align}
	D_t \partial_{\dot{q}} L - \partial_q L\label{e:EL}
	 & = 0
\end{align}
In the case where one or both of the end-points are free (i.e., there is no boundary condition for the path, and thus no constraint on the variation), then the first term on the right-hand side of Eq.~\eqref{e:dS} provides an additional set of natural boundary conditions that must be satisfied,
\begin{align}
	\left.\partial_{\dot{q}} L\right\rvert_{t_1,t_2} & = 0.\label{e:natbnd}
\end{align}

\subsection{Optimal Control}\label{s:theory_opt}
We now consider a specialized version of the Euler-Lagrange equations, adapted to typical optimal control problems, as encountered in finance and biology\cite{Lenhart2007,Lambertini2018}.
We are interested in understanding how an ``agent'' (e.g., an individual), behaves in response to its environment (e.g., the population). Let the state of the population and the individual be specified by $\theta$ and $\psi$, respectively, with $k$ and $\kappa$ the corresponding control variables (encoding behaviour), which are time-dependent, changing in time in response to $\theta$ and $\psi$. We will assume that the dynamics of the population is known $\dot{\theta} = F$, and furthermore, we assume that the population is composed of identical individuals, such that $F$ also determines the individual dynamics. However, we note that $F$ should distinguish between population and individual state variables. Without loss of generality, we consider dynamical equations of the form,
\begin{align}
	\dot{\theta} & = F(t, \theta, \theta, k) \label{e:dottheta_gen}  \\
	\dot{\psi}   & = F(t, \theta, \psi, \kappa).\label{e:dotpsi_gen}
\end{align}
from which we see that $\dot{\theta}=\dot{\psi}$ whenever $\theta=\psi$ and $k=\kappa$ (i.e., at the Nash equilibrium).

Individuals should behave in order to maximize their total ``utility'' (playing the role of the action).  While the dynamical equations are fixed, $\dot{\psi} = F$, the time evolution can be actively controlled by the individual through their choice of $\kappa$.
The functional we wish to extremize, analogous to Eq.~\eqref{e:S}, is the integrated instantaneous payoff per unit time $V$, with respect to the control $\kappa$,
\begin{align}
	\delta_\kappa\int_{t_1}^{t_2}         & V(t, \theta(t), \psi(t), \kappa(t))\dd{t} = 0 \label{e:optctrl} \\
	\mathrm{subject\, to\quad} \dot{\psi} & = F(t, \theta(t), \psi(t), \kappa(t)), \notag                   \\
	\psi(0)                               & = \psi_\mathrm{init}\notag.
\end{align}
Note that, at this point, the population state variables are considered as external fields, i.e., they are passive variables with regards to the variations.
We have assumed that the payoff function $V$ (playing the role of the Lagrangian) depends only on time and coordinates, but not on the generalized velocities, such as $\dot{\psi}$ or $\dot{\kappa}$. Thus, $V$ could be considered as the potential energy contribution to a Lagrangian with no kinetic energy term. However, this does not mean that no velocity dependence is possible. Thanks to the dynamical constraints of Eqs.~(\ref{e:dottheta_gen}-\ref{e:dotpsi_gen}), we can always express $\dot{\theta}$ or $\dot{\psi}$ in terms of the ``coordinates'' $\theta$, $\psi$, $k$, $\kappa$. This only precludes terms in $\dot{\kappa}$. Furthermore, since the constraint is integrable or holonomic, this constrained optimization problem can be written in terms of an unconstrained optimization, by introducing an augmented Lagrangian $L$, with additional degrees of freedom (corresponding to the Lagrange multipliers $\lambda$)\cite{sicm}
\begin{align}
	L(t; \theta, \psi, \kappa, \lambda; \cdot, \dot{\psi},\cdot,\cdot)
	 & = V(t,\theta, \psi, \kappa) \label{e:L0}                                        \\
	 & \quad + \lambda\cdot\left(F(t, \theta, \psi, \kappa) - \dot{\psi}\right) \notag \\
	 & = L^\prime(t,\theta, \psi,\kappa,\lambda) - \lambda\cdot\dot\psi\label{e:L}.
\end{align}
For what follows we have expressed this Lagrangian in terms of an auxiliary function $L^\prime$,
\begin{align}
	L^\prime(t,\theta, \psi, \kappa, \lambda) & = V(t,\theta,\psi,\kappa) + \lambda\cdot F(t,\theta,\psi,\kappa)\label{e:Lprime}
\end{align}
which does not depend explicitly on the velocities. We use colons to explicitly divide time, coordinate, and velocity variables in the Lagrangian, and its derived quantities, in cases where multiple components are used, and we mark independence with respect to a given component by writing an empty slot $(\cdot)$ into the corresponding function argument. Note that, by construction, the velocity dependence in $L$ is linear, and it is due solely to the constraint term $\lambda~\cdot~\dot{\psi}$.

We can abstract away the individual level variables in terms of generalized coordinates and velocities, $q$ and $\dot{q}$,
where the individual components, $\psi$ and $\lambda$, can themselves be $n$-dimensional vectors, though we only consider a single scalar control variable $\kappa$
\begin{align}
	q & =\begin{pmatrix}\psi   \\
		     \kappa \\
		     \lambda
	     \end{pmatrix},\quad
	\dot{q} =\begin{pmatrix}\dot\psi   \\
		         \dot\kappa \\
		         \dot\lambda
	         \end{pmatrix}.\label{e:q}
\end{align}
The Euler-Lagrange equations for this Lagrangian $L(t;\theta,q;\cdot,\dot{q})$, obtained by extremizing the utility with respect to variations in the individual degrees of freedom $q$ (with $\theta$ fixed), are (Eq.~\ref{e:EL})
\begin{align}
	D_t \begin{pmatrix}
		    -\lambda \\
		    0        \\
		    0
	    \end{pmatrix} & =
	\begin{pmatrix}
		-\dot{\lambda} \\
		0              \\
		0
	\end{pmatrix}=
	\begin{pmatrix}
		\partial_\psi L^\prime   \\
		\partial_\kappa L^\prime \\
		\partial_\lambda L
	\end{pmatrix}.\label{e:ELopt}
\end{align}
The first equation in~\eqref{e:ELopt} determines the dynamical equation for the Lagrange multipliers
\begin{align}
	\dot{\lambda}(t)
	 & = -(\partial_{\psi}L^\prime)(t, \theta, \psi, \kappa_\opt, \lambda)\label{e:dotlambda}
\end{align}
which should be evaluated at the optimal control $\kappa_\opt$, obtained from the optimality condition defined by the second equation
\begin{align}
	(\partial_\kappa L^\prime)(t,\theta,\psi,\kappa,\lambda)\big\rvert_{\kappa=\kappa_\opt} & = 0,\label{e:optimality}
\end{align}
which provides an implicit definition for $\kappa_\opt$ in terms of the state variables
\begin{align}
	\kappa_\opt & \equiv \kappa_\opt(t, \theta, \psi, \lambda).\label{e:kopt}
\end{align}
The last of these equations simple reproduces the constraint, since $\partial_\lambda L = F - \dot{\psi}$,
\begin{align}
	\dot\psi & = F(t, \theta,\psi, \kappa_\opt).\label{e:dotpsi}
\end{align}
Eqs.~\eqref{e:dotlambda}~and~\eqref{e:dotpsi} determine a set of $2n$ first order differential equations, requiring $2n$ boundary conditions.
We are assuming that the initial condition for $\psi$ is given (i.e., $\psi(t_1) = \psi_0$); the remaining $n$ conditions are provided by the open boundary conditions at $t_2$,
\begin{align}
	\left.\partial_{\dot{q}} L\right\rvert_{t_2} & = \left.\partial_{\dot{\psi}} L\right\rvert_{t_2} = -\lambda(t_2) = 0.\label{e:lambdabnd}
\end{align}
Because of the specific velocity dependence in $L$, the dynamical equations for $\lambda$ and the optimality condition are determined uniquely by the $L^\prime$ function, which has no velocity dependence. In fact, this $L^\prime$ can be shown to be equivalent to (minus) the corresponding Hamiltonian, with $\lambda$ taking on the role of momenta conjugate to $\psi$, since $\partial_{\dot{q}} L = \partial_{\dot{\psi}} L = -\lambda$. In this case, not only are Lagrangian and Hamiltonian formulations equivalent, but they result in exactly the same set of $2n$ first-order differential equations for $\psi$ and $\lambda$ (see Appendix~\ref{s:hamilton}).

\subsection{Optimal Decision Making during Epidemics}\label{s:theory_epi}
We will now consider the problem of determining optimal social distancing during an epidemic like the current SARS-CoV-2 pandemic.
For simplicity, we will assume that the epidemic follows SIR dynamics\cite{Petard1938}, described by the fraction of the population that is susceptible $s$, infectious $i$, and infected $r$ as a function of time. Since $r$ is slaved to $i$ in this representation, we need not explicitly solve for it. The state of the population $\theta$, and its dynamics, is then given by
\begin{align}
	\theta       & = \begin{pmatrix}s \\ i \end{pmatrix}
	\label{e:theta_sir}                                                                      \\
	\dot{\theta} & = \begin{pmatrix}- k s i \\ k s i - i\end{pmatrix},\label{e:dottheta_sir}
\end{align}
with $k$ the population average level of infectiousness, which we use as a proxy to describe the population behaviour or strategy.
Against this backdrop, we now consider an individual that is capable of adopting a different strategy $\kappa$. Let $\psi_s$ and $\psi_i$ denote the probability that the individual is susceptible or infectious, respectively. The individual state dynamics are given by\cite{Reluga2010}
\begin{align}
	\psi       & = \begin{pmatrix}
		               \psi_s \\ \psi_i
	               \end{pmatrix} \label{e:psi_sir} \\
	\dot{\psi} & \equiv F(\theta, \psi, \kappa) =
	\begin{pmatrix}
		-\kappa \psi_s i \\
		\kappa \psi_s i - \psi_i
	\end{pmatrix}.\label{e:dotpsi_sir}
\end{align}
The individual(s), which we assume to be rational with access to perfect information, will choose their strategy to optimize their total utility, obtained by integrating their instantaneous payoff $V$. For simplicity, we adopt the following form for $V$\cite{Schnyder2022}
\begin{align}
	V(\theta, \psi, \kappa) & = -\alpha(i) \psi_i - \beta(\kappa - \kappa^\star)^2.\label{e:payoff_sir}
\end{align}
Where, without loss of generality, we have assumed that $V$ does not depend explicitly on time (e.g., there is no discounting), nor on the population strategy $k$.
The first term on the rhs of Eq.~\eqref{e:payoff_sir} represents the cost of being infected, which can depend on $i$, to account for health-care thresholds, while the second term represents the cost of reducing social activity, with respect to the preferred or natural state $\kappa^\star$. This quantity is also known as the basic reproduction number $R_0$. The optimal strategy is given by Eqs.~\eqref{e:dotlambda}-\eqref{e:dotpsi}. The optimality condition is obtained from the partial derivatives of the Lagrangian with respect to the control parameter
\begin{align}
	(\partial_\kappa L^\prime)(\theta,\psi,\kappa,\lambda)
	 & = (\partial_\kappa V + \lambda\cdot\partial_\kappa F)(\theta,\psi,\kappa) \label{e:optimality_sir0} \\
	 & = -2\beta(\kappa - \kappa^\star) - \psi_s i (\lambda_s - \lambda_i)\label{e:optimality_sir}
\end{align}
from which we obtain the following closed form solution for $\kappa_\opt$
\begin{align}
	\kappa_\opt(\theta,\psi,\lambda) & = \kappa^\star - \frac{1}{2\beta}\psi_s i (\lambda_s - \lambda_i).\label{e:kopt_sir}
\end{align}
The dynamics of the Lagrange multipliers are given by
\begin{align}
	\dot{\lambda}      & = -(\partial_{\psi} V + \lambda\cdot\partial_{\psi} F)(\theta,\psi,\kappa_\opt) \label{e:dotlambda_sir0} \\
	\begin{pmatrix}
		\dot{\lambda}_s \\
		\dot{\lambda}_i
	\end{pmatrix} & = \begin{pmatrix}
		                  \kappa_\opt i (\lambda_s - \lambda_i) \\
		                  \alpha(i) + \lambda_i
	                  \end{pmatrix}.\label{e:dotlambda_sir}
\end{align}
We are interested in the Nash equilibrium solution, for which individuals all adopt the same strategy $k=\kappa=\kappa_\opt$, i.e.,  there is no benefit in assuming an alternative (defector) strategy. In this case, not only is the population and individual strategy equal, but the states of the population and individual should also be equivalent (i.e., $s=\psi_s$, $i = \psi_i$). This type of approach has been used extensively to study optimal social distancing\cite{Reluga2010,Wang2016,McAdams2020,Makris2020}, as well as optimal government intervention strategies\cite{Toxvaerd2019,Rowthorn2020,Bethune2020,Eichenbaum2021,Schnyder2022}.

\section{Nash Neural Networks}\label{s:n3}
\begin{figure*}[ht!]
	\centering
	\includegraphics[width=0.95\textwidth]{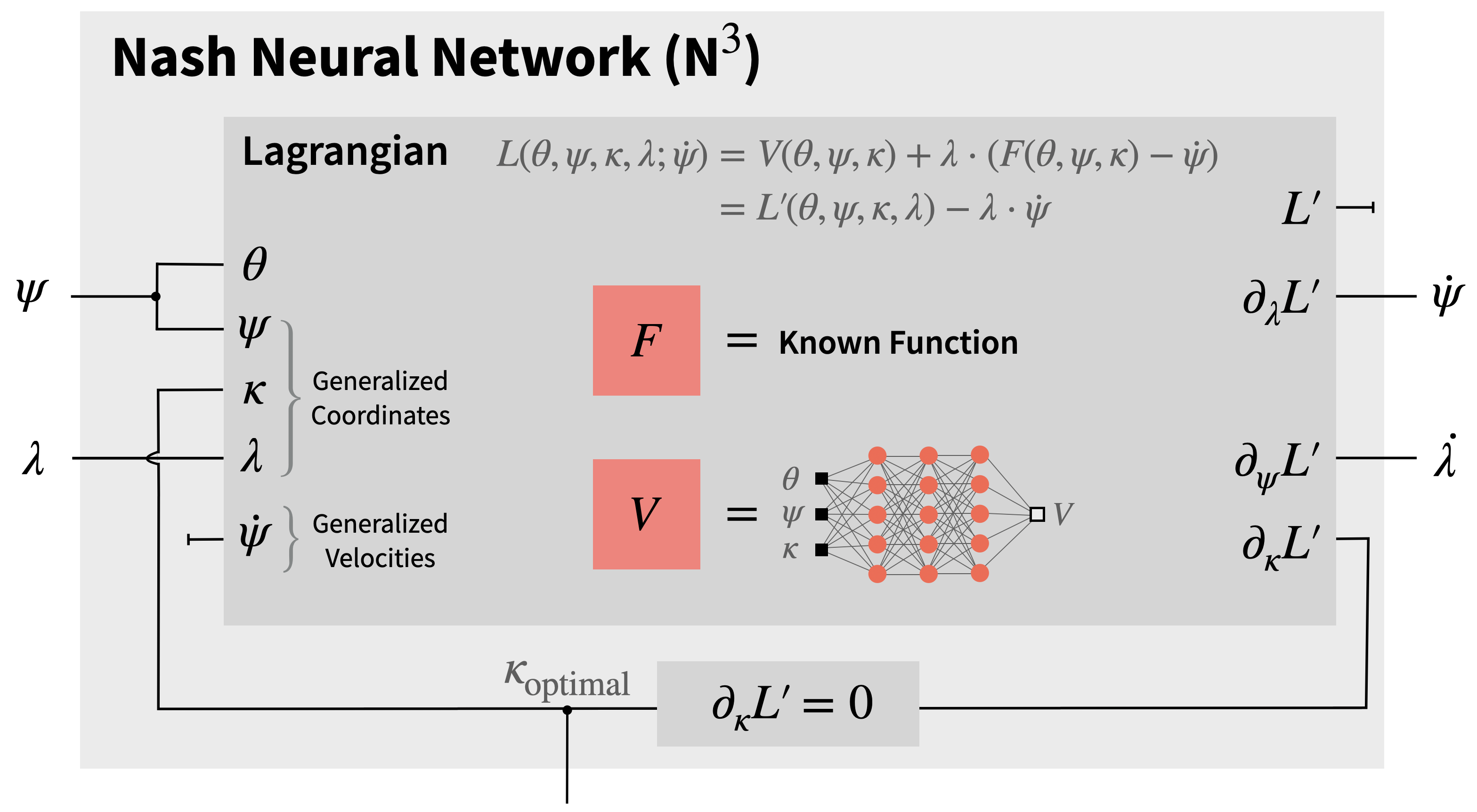}
	\caption{\label{f:n3}A schematic ``wiring diagram'' describing the relationship between the known constraint function $F$, the black-box payoff function $V$, the augmented Lagrangian $L$, and the system derivatives used to define the Euler-Lagrange equations for the corresponding optimal control problem. Note that, while the Lagrangian is formally a function of the generalized coordinates $q$ and velocities $\dot{q}$, we only need to supply the coordinates corresponding to the state of the individual $\psi$, and the Lagrange multipliers. By construction, the optimality condition is self-consistently solved for, providing the appropriate value of $\kappa=\kappa_\opt$ to use as input, and the required system derivatives have lost the functional dependence on the velocities. Furthermore, at the Nash equilibrium, individual and population states are equivalent $\theta=\psi$, which is why only a single (branching) input channel is required, though the Lagrangian itself must differentiate between the two. Thus, this $N^3$ provides a state evolver, allowing us to compute the time-rate of change of $\psi$ ($\theta$) and $\lambda$, as well as the optimal control variable $\kappa_\opt$, as a function of $\psi$ and $\lambda$ only.}
\end{figure*}

Let us now consider the ``inverse'' optimal control problem, that of inferring the individual payoff function $V$, from observations of the pandemic behavior. For this task, we will build upon the Hamiltonian\cite{Greydanus2019} and Lagrangian\cite{Cranmer2020} Neural Networks, to develop a novel Physics Informed Neural Network (PINN), capable of self-consistently solving the optimality problem for these types of differential games with Nash equilibria.
We note that, while we have assumed that $F$ is known, it could also be considered as an additional unknown function to be learned. Thus, for the epidemics example, we could consider to include $F$ as a disease informed neural network\cite{Shaier2021}.
Our neural network, which we refer to as a Nash Neural Network ($N^3$), is constructed in such a way that it respects the Euler-Lagrange equations of the underlying optimal control problem (as defined by the known $F$ and the black-box payoff function $V$), and is able to self-consistently compute the optimal control and evaluate itself at the Nash equilibrium. A schematic representation of the proposed network is given in Fig.~\ref{f:n3}. We start with a hidden network used to represent the unknown payoff function $V$. All we assume regarding this black-box function is its signature or functional dependence, $V(\theta,\psi,\kappa)$. This function is then combined with the dynamical constraint $F$ (i.e., SIR model) in order to construct the augmented Lagrangian $L$, and more specifically $L^\prime$, which introduces the Lagrange multipliers $\lambda$ as additional coordinates. Then, we take the appropriate system derivatives, leveraging automatic differentiation capabilities, in order to construct the Euler-Lagrange equations for this optimal control problem. The outputs of this procedure are three secondary neural networks (derived from the neural network encoding $V$) that allow us to compute $\partial_\kappa L^\prime$, $\partial_\psi L^\prime$, and $\partial_\lambda L^\prime$, the optimality condition, and $\lambda$ and $\psi$ dynamics of Eqs.\eqref{e:dotlambda}-\eqref{e:dotpsi}, respectively. Furthermore, we numerically solve the optimality condition for $\kappa_\opt$, and use this value as input to (all) the networks. Thus, the $N^3$ evaluates itself on the self-consistently determined optimal control parameter. This is the main difference between our approach and that of the PDP\cite{Jin2020a,Jin2020b}. We are learning the Lagrangian/Hamiltonian, not just the payoff function, and directly including the optimality condition as an additional bias in the network structure. Finally, we only ever evaluate the network at the Nash equilibrium, at which point $\theta=\psi$, $k=\kappa=\kappa_\opt$, but the distinction between population and individual state variables is maintained throughout, as this is necessary to derive the appropriate Euler-Lagrange equations. In order to be able to train the network, we must be able to compute gradients with respect to the network parameters through the optimization problem ($\kappa_\opt$). This is accomplished using implicit automatic differentiation\cite{Margossian2021,Blondel2021}.

For what follows, when we apply the $N^3$ to study the Nash equilibria governing the optimal behaviour during an epidemic, it helps to analyze where the different terms in Eqs.~(\ref{e:optimality_sir0}~-~\ref{e:dotlambda_sir}), come from. Both the optimality condition and the $\lambda$ dynamics contain two terms, one derived from the payoff function $V$, and the other from the dynamical constraints $F$. When performing the learning, $V$ will be an unknown black-box function that we want to learn, while $F$ is a known function (i.e., SIR dynamics). Consider the optimality conditions of Eq.\eqref{e:optimality_sir0}, the second term on the rhs is known exactly (as a function of $\theta$, $\psi$, $\kappa$, and $\lambda$), which leaves only the first term to be learned. For the particular case we are considering here, this remaining term $\partial_\kappa V$, contains no $\theta$, $\psi$, or $\lambda$ dependence; it depends only on $\kappa$. This means that we should be able to learn the $\kappa$ dependence of the payoff function ($-\beta(\kappa-\kappa^\star)^2$) solely from observations of the optimal behaviour $\kappa_\opt$. Considering the $\lambda$ dynamics, we see that $k_\opt$ should also be enough to learn $\dot{\lambda}_s$, since $V$ is independent of $\psi_s$ ($\partial_{\psi_s}V =0$), which means that $\dot{\lambda}_s$ is uniquely determined by $F$. However, the same does not apply to $\dot{\lambda}_i$, due to the cost of being infected ($\propto \alpha(i) \psi_i$), which depends explicitly on $\psi_i$. Thus, to fully learn the $\dot{\lambda}$ dynamics, we would need to also include measurements of the dynamics into the training data. While this analysis is based on the specific form of the payoff function we are using to generate the training data, our conclusions can be generalized to other payoffs. First, we will require knowledge of both the population/individual state variables, the optimal control, and the Lagrange multipliers in order to infer the functional dependence of an arbitrary payoff function. Second, in the special case that the payoff contains a term that is independent of $\psi$, we can expect to learn it only from observations of $\kappa_\opt$. Finally, any term in the payoff function that depends only on $\theta$ cannot be learned, since it would have no effect on the optimality condition or the system dynamics (see Appendix~\ref{s:uniqueness}).

For the learning, we start by assuming that we have complete knowledge of the system dynamics, this includes the population/individual state variables $\theta=\psi$, the Lagrange multipliers $\lambda$, as well as their time derivatives $\dot{\psi}$ and $\dot{\lambda}$, and the optimal control $\kappa_\opt$.
In practice we would just require $\theta(t)$ ($\lambda(t)$), since this would allow us to compute $\dot{\psi}$ ($\dot{\lambda}$), as well as $\kappa_\opt$. With this information, we are in a position to train our $N^3$, as it is constructed to predict $(\dot{\theta},\dot{\lambda})$ as a function of $(\theta,\lambda)$, evaluated at the Nash equilibrium for $k=\kappa=\kappa_\opt$. We will consider the following two loss functions, $\op{L}_\kappa$ and $\op{L}_{\kappa,\lambda}$, defined as
\begin{align}
	\op{L}_\kappa           & = \avg{\delta\kappa^2} + \sum_j (\partial_\kappa L^\prime)^2 \label{e:loss_k} \\
	\op{L}_{\kappa,\lambda} & = \avg{\delta\lambda^2} + \op{L}_\kappa \label{e:loss_kl}                     \\
	\avg{\delta X^2}        & = \sum_j \left(X^{\nn} - X^{\train}\right)^2\label{e:avg}
\end{align}
where $\avg{\delta X^2}$ measures the mean squared error between the training data and the neural network prediction, and the sum is over all training points $j$. The first loss function, $\op{L}_\kappa$, trains exclusively on the optimal control. As such, it should learn both the $\kappa$ and $\psi$ dynamics, which don't depend on $\lambda$, by definition. Furthermore, given the particular form of the payoff function we are using, it will also learn the social distancing term and the $\lambda_s$ dynamics. The second loss function, $\op{L}_{\kappa,\lambda}$ trains on the optimal control and the $\lambda$ dynamics. This will allow us to learn the full functional dependence of the payoff function, i.e., the remaining cost of infection term. Note that we have included an additional term in both loss functions, given by the sum of squares of the optimality condition $\sum_j (\partial_\kappa L^\prime)^2$. This was done for technical reasons\footnote{At the time of writing, the JAXopt\cite{Blondel2021} library only supported root-finding using a simple bisection algorithm, which required specifying a bracketing interval around the root. This made it impractical to use during training.}, as we have replaced the root finder with a minimizer. In this way, we have decomposed the task of finding the roots of the optimality condition into a nested minimization procedure: the $N^3$ will produce a $k_\opt$ that is a (local) minimum of the optimality condition, but during the training, as the loss function is minimized, this minimum is itself minimized. This is similar in spirit to standard multidimensional root-finding algorithms, such as Powell's hybrid method, in which Newton and gradient direction steps are interleaved.
\section{Results}
\begin{table}[ht!]
	\begin{tabular}{C | C | C}
		\alpha_0 & \op{L}_{\kappa}                         & \op{L}_{\kappa,\lambda}               \\
		\hline
		100      & 3.2\times 10^{-6}   (3.3\times 10^{-6}) & 1.1\times 10^{-5} (1.5\times 10^{-5}) \\
		200      & 3.2\times 10^{-9} (1.0\times 10^{-6})   & 7.9\times 10^{-5} (2.4\times 10^{-4}) \\
		400      & 1.7\times 10^{-6} (1.8\times 10^{-6})   & 1.2\times 10^{-3} (9.6\times 10^{-4})
	\end{tabular}
	\caption{\label{t:loss} Loss functions for the three different training sets, using the $\op{L}_{\kappa}$ and $\op{L}_{\kappa,\lambda}$ loss functions. Note that, for equal number of steps, training against both the optimal control $\kappa_\opt$ and the $\lambda$ dynamics results in a loss function that can be up to three orders of magnitude higher. Results in parenthesis show the loss evaluated on the test points, not included in the training.}
\end{table}
\begin{figure}[ht!]
	\centering
	\includegraphics[width=\columnwidth]{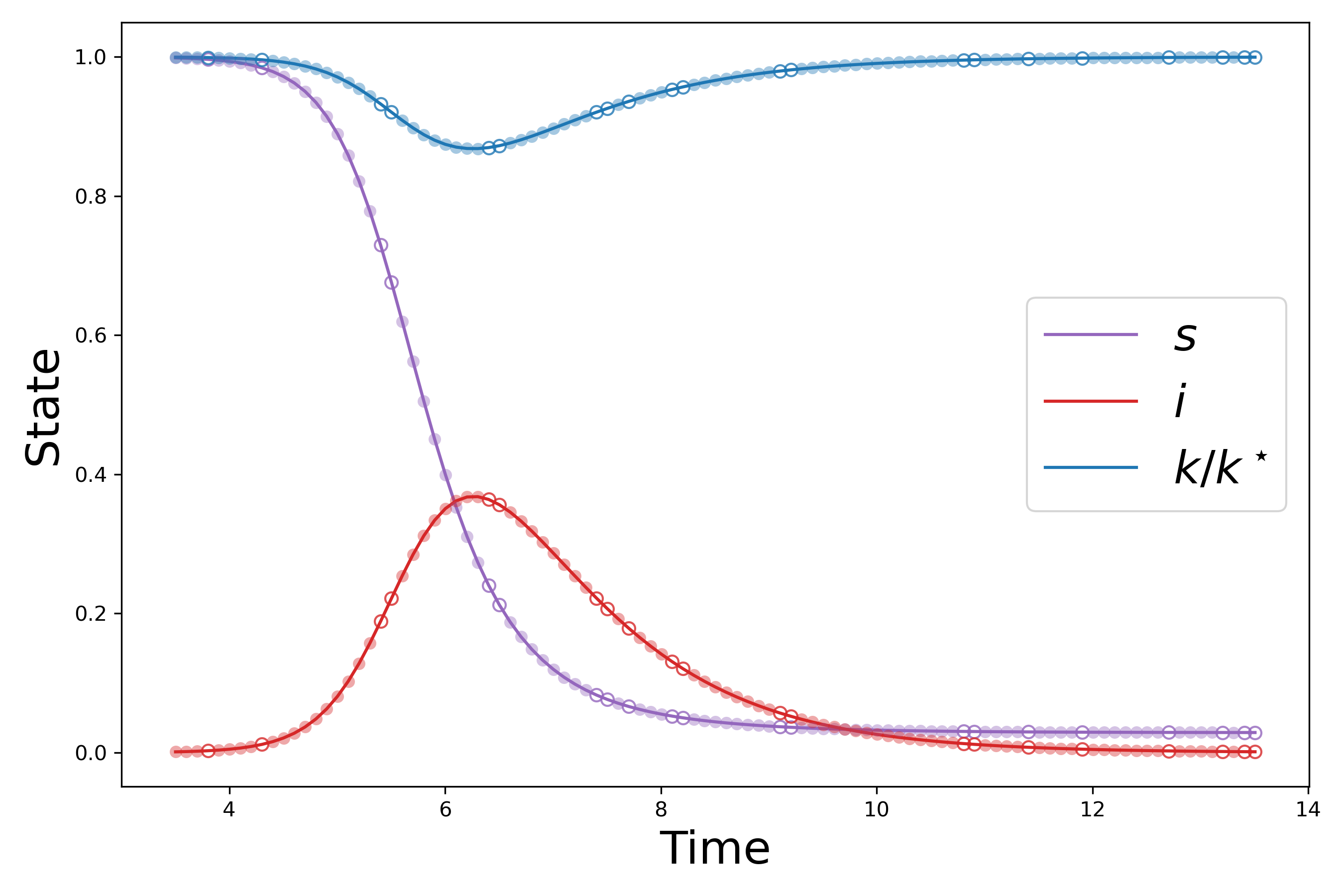}\\
	\includegraphics[width=\columnwidth]{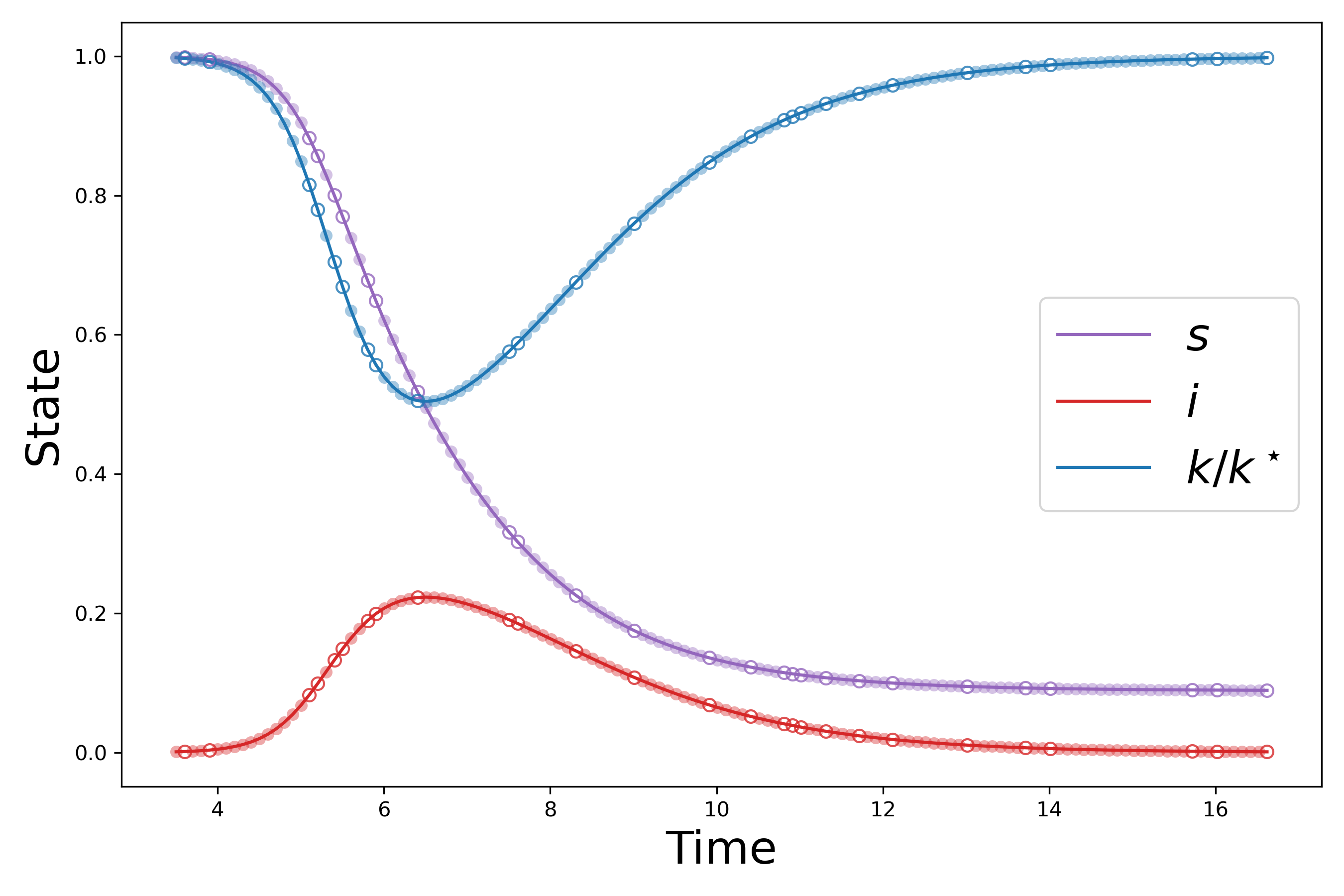}\\
	\includegraphics[width=\columnwidth]{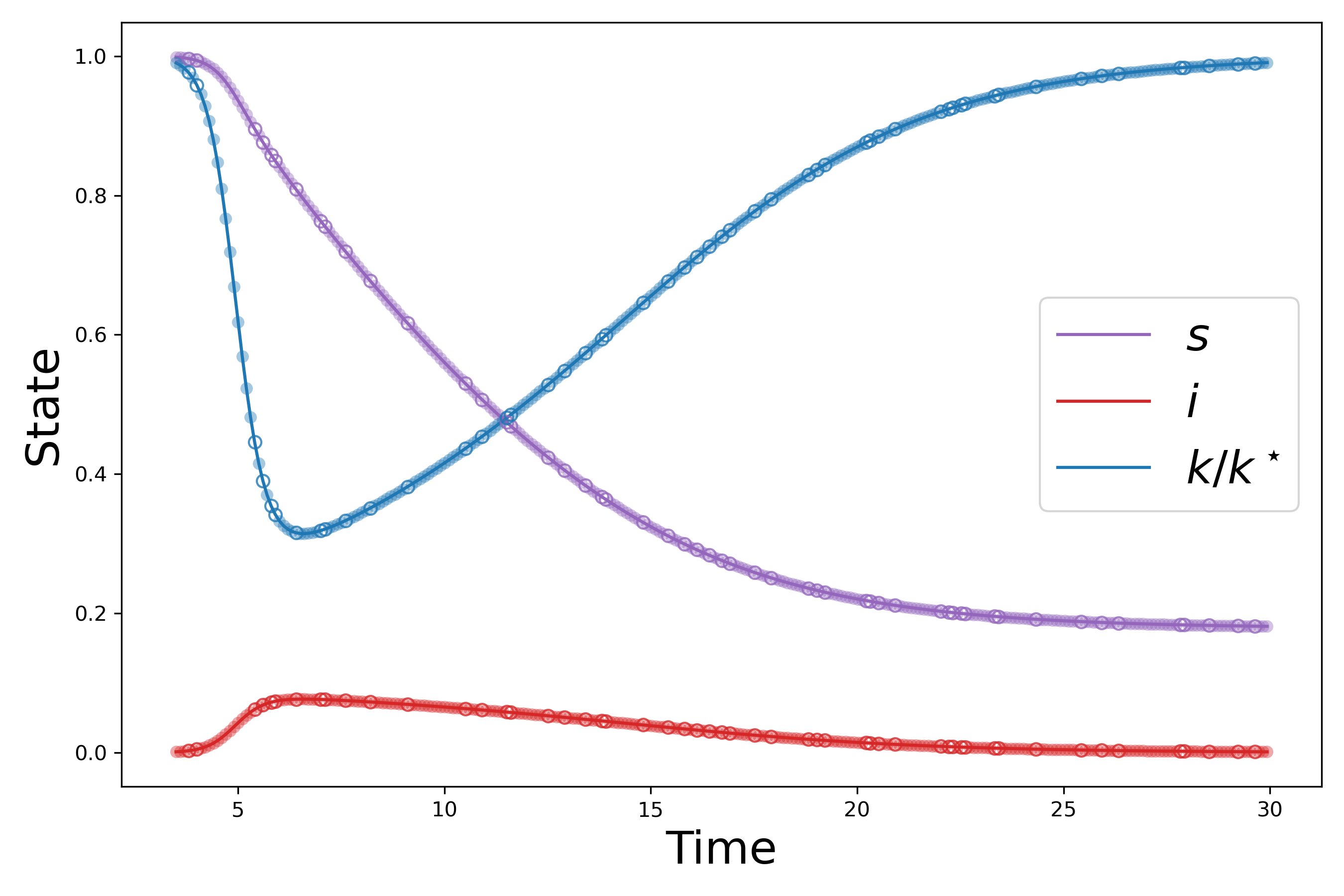}
	\caption{\label{f:sir} Nash solutions to the optimal distancing policy for different infection cost $\alpha_0$, (top) $100$, (middle) $200$, and (bottom) $400$, resulting in weak, moderate, and strong social distancing, respectively\cite{Schnyder2022}. The filled symbols correspond to the training data points, the empty symbols to the test point, used to check against over-fitting. Note that we have only used the portion of the trajectory for which $i \ge 10^{-3}$, which contains the interesting part of the dynamics, i.e., $\dot{\psi}$ and $\dot{\lambda}$ do not vanish.}
\end{figure}
We use synthetic data generated by solving Eqs.~\eqref{e:dotpsi_sir}~and~\eqref{e:dotlambda_sir}, under Nash equilibrium conditions, for which $\theta=\psi$ and $k=\kappa=\kappa_\opt$, with the optimal control determined by Eq.~\eqref{e:kopt_sir}. The boundary conditions for $\theta$ and $\lambda$, at the initial $t_0=0$ and final times $t_f$, respectively, were $\theta(0) = (s(0),i(0))= (1-i_0, i_0=10^{-8})$, and $\lambda(t_f) = (\lambda_s(t_f), \lambda_i(t_f)) = (s_f, i_f)$ (corresponding to vaccination boundary conditions). The equations were solved using an iterative forward-backward sweep method, until convergence of $\kappa$ was obtained, with final times $t_f\simeq 100$. In total, we have considered three different (constant) values of $\alpha(i) = \alpha_0 = 100, 200, 400$, with $\beta = 1$, and $\kappa^\star = 4$. A full description of the optimal behaviour, with and without governemnt intervention is provided in Ref.\cite{Schnyder2022}. The Nash solutions, used as training data, are shown in Fig.~\ref{f:sir} for the three values of $\alpha_0$, corresponding to weak, moderate, and strong social distancing. For the learning, we used the ADAM optimizer\cite{Kingma2015}, with a step size of $5\times 10^{-5}$, and default values for $b_1 = 0.9$, $b_2=0.999$, and $\epsilon=10^{-8}$. Unless otherwise stated, results are obtained using a neural network with three hidden layers of $N=200$ neurons each, with hyperbolic-tangent activation functions. Network parameters were initialized using a standard normal distribution, with mean $0$ and variance $1/N$. Training was stopped after $1.2\times 10^5$ steps.

To account for the fact that the Lagrangian, and thus the payoff function, is not unique, we will only plot differences in the payoff function (see Appendix\ref{s:uniqueness}). In contrast to canonical physical systems, where the potential energy can be offset by a constant, here the payoff function can be offset by a function of time and population state variables. Therefore, when analyzing the $\kappa$ dependance in $V$, we will consider $V(\theta, \psi, \kappa)- V(\theta, \psi, \kappa^\star)$, whereas for the $\theta$/$\psi$ dependence we consider $V(\theta, \psi, \kappa) - V(\theta, 0, \kappa)$. Notice that for the ``real'' utility used to generate the training data, these shifted utilities would exactly provide the two terms in the payoff, $-\beta(\kappa - \kappa^\star)^2$ and $-\alpha(i)\psi_i$, respectively.
\subsection{Learning from Behaviour}
\begin{figure}[ht!]
	\centering
	\includegraphics[width=\columnwidth]{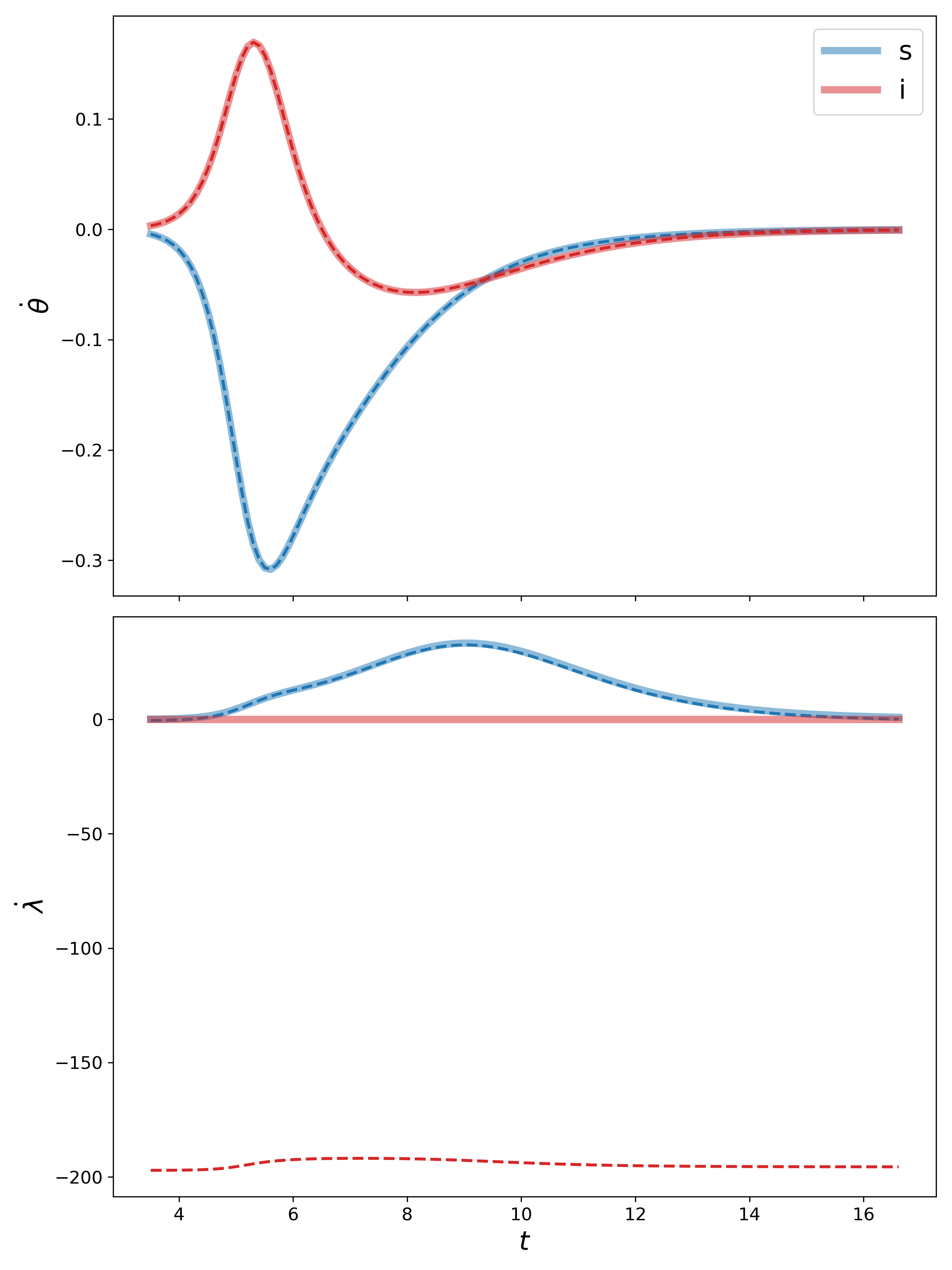}
	\caption{\label{f:d1_l0_dstate} Comparison between the exact (solid) solution and the (dashed) $N^3$ predictions, trained using $\op{L}_\kappa$, for the state dynamics, $\dot{\theta}$ and $\dot{\lambda}$. The results are computed on the $\alpha_0=200$ pandemic trajectory shown in the middle panel of Fig.\ref{f:sir}, plotted as a function of time. The corresponding $\lambda$ values, also required as input to the neural network are not shown.}
\end{figure}
The results obtained from learning only against the optimal control $\kappa_\opt$ provide excellent predictions for $\kappa$, $\dot{\theta}$, and $\dot{\lambda}_s$, as expected. This is seen in Fig.~\ref{f:d1_l0_dstate} for the case of $\alpha=200$, where we plot the exact solution, together with the $N^3$ predictions, as a function of $\theta$ and $\lambda$. While we did not train against $\dot{\lambda}$ explicitly, the fact that the $\dot{\lambda}_s$ is uniquely determined by the constraint $F$ and the social distancing ($\beta$) term in the payoff function, means that it can be completely recovered from $\kappa_\opt$, since the latter is also determined from these two terms. The dynamics for $\lambda_i$ is clearly ``wrong'', offset by what seems to be a constant $\simeq \alpha_0$. In fact, the predictions shown here correspond to the $F$ contribution to $\dot{\lambda}_i$, equal to $-\lambda\cdot\partial_{\psi_i} F = \lambda_i$, as shown in Eq.~\eqref{e:dotlambda_sir}. The remaining contribution, coming from the payoff term $-\partial_{\psi_i} V = \alpha(i) = \alpha_0$, cannot be learned from $\kappa_\opt$, as it does not couple directly to the control variable. For the same reason, we are unable to predict the $i$ dependance of $V$ (not shown).

\begin{figure}[ht!]
	\centering
	\includegraphics[width=\columnwidth]{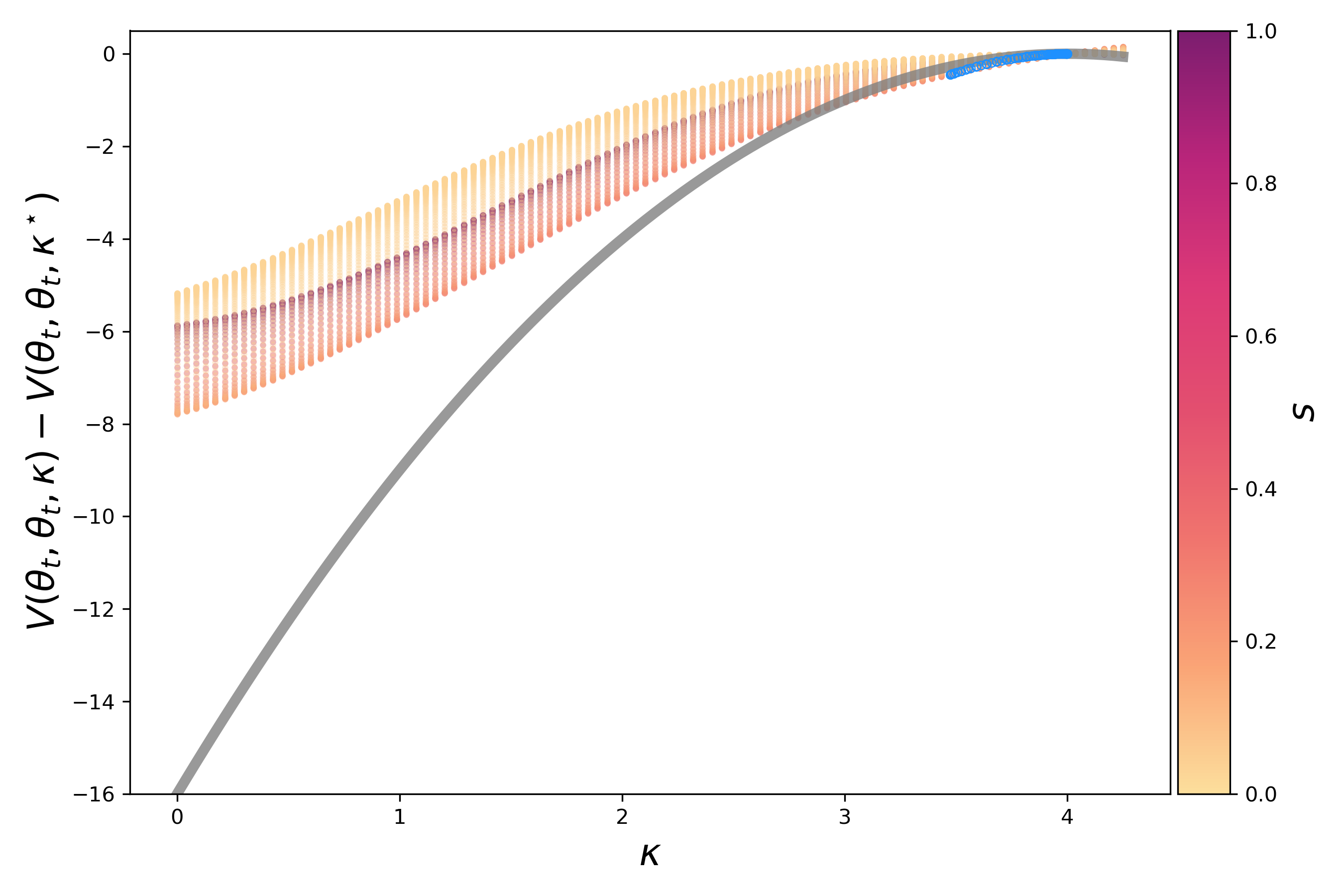}
	\includegraphics[width=\columnwidth]{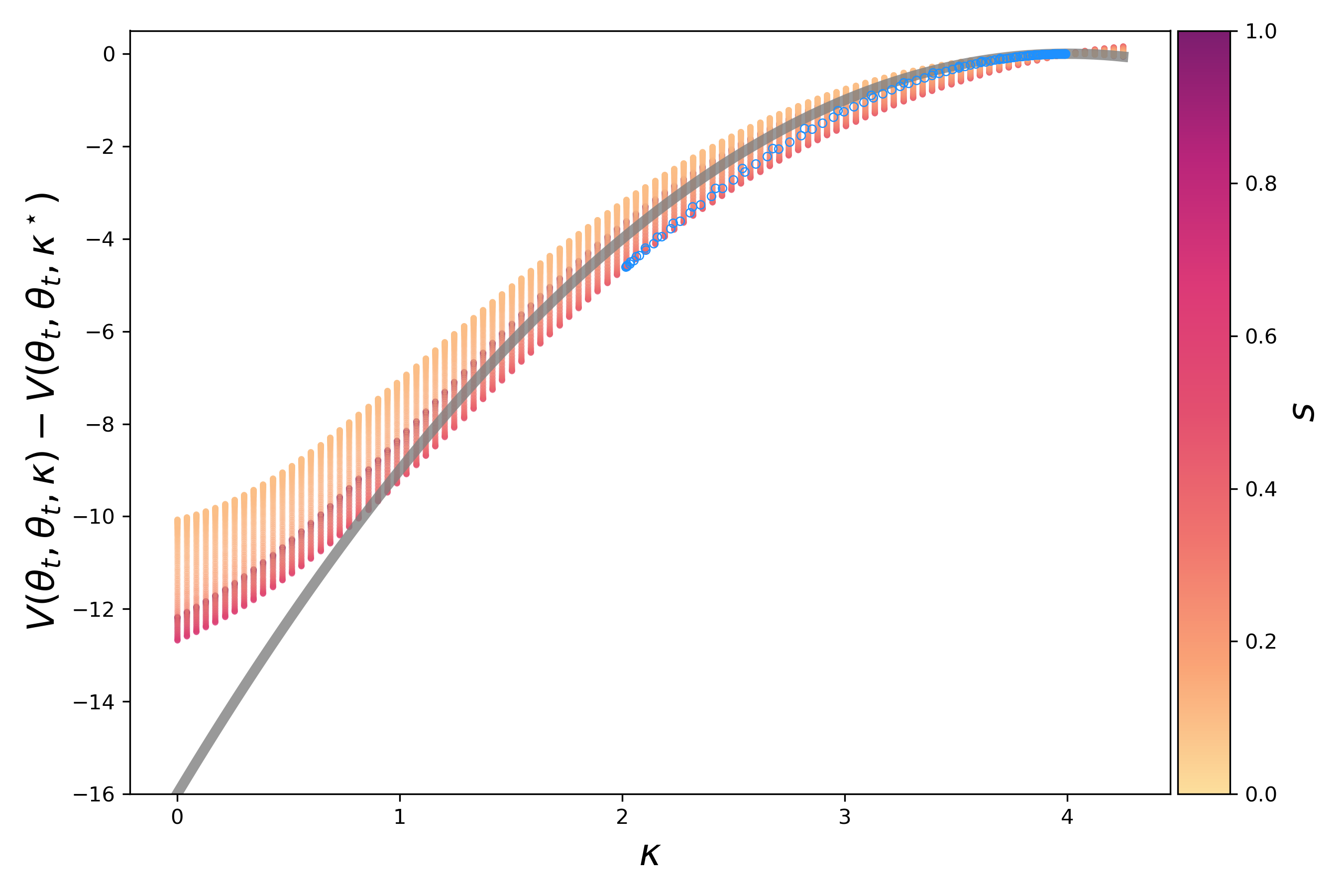}
	\includegraphics[width=\columnwidth]{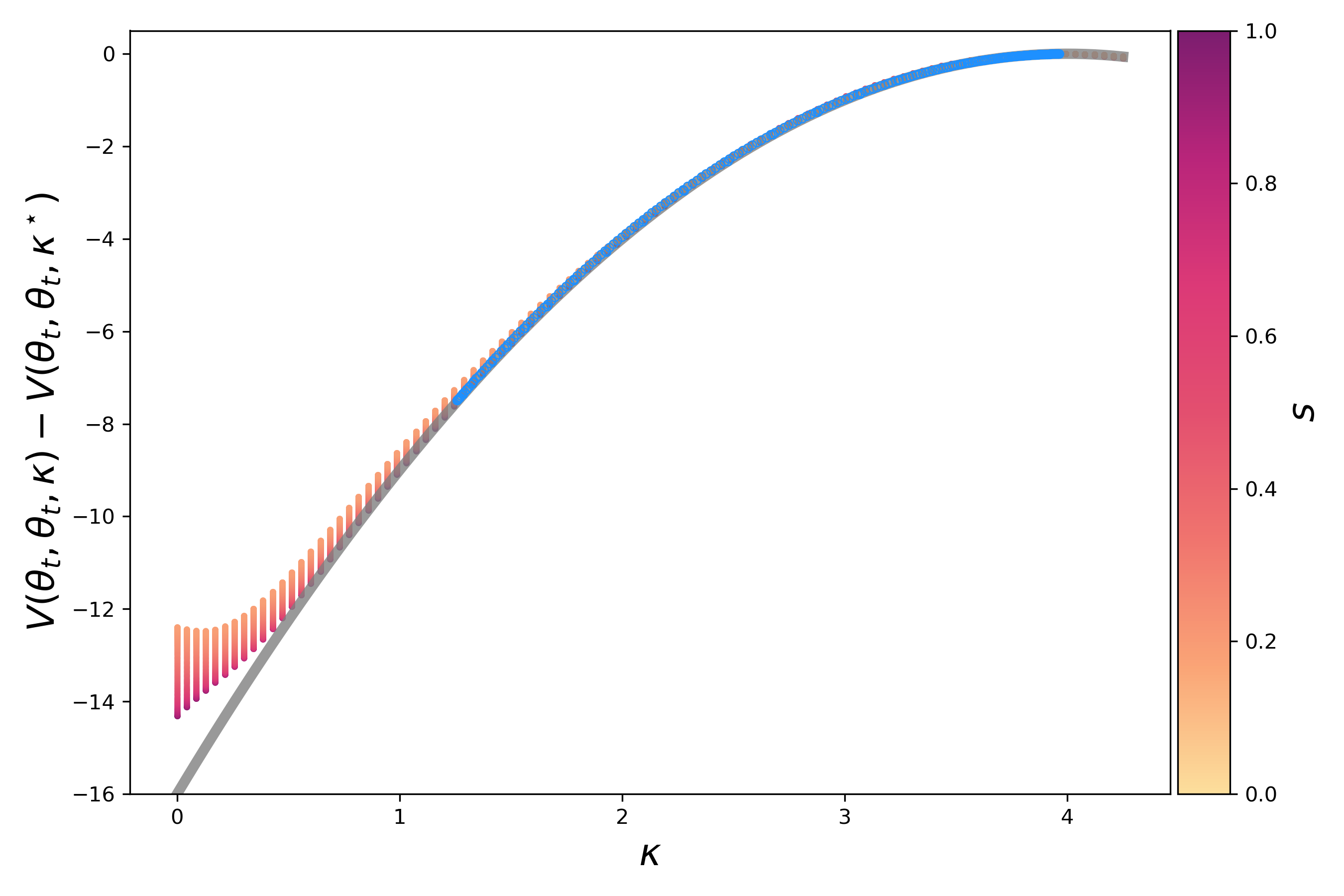}
	\caption{\label{f:l0_uk}$N^3$ predictions for the payoff $V$, as a function of $\kappa$, evaluated at distinct points along the pandemic trajectory $\theta_t = (s_t, i_t)$, for the three values of $\alpha$, (top) $100$, (middle) $200$, and (bottom) $400$. Color encodes the corresponding value of time (susceptibles), and the solid lines gives the exact theoretical value $-\beta(\kappa-\kappa^\star)^2$. The empty blue circles show the predictions on the pandemic/training data set, i.e, using the optimal behaviour $\kappa_\opt$, such that there is one datum per curve.}
\end{figure}
Although $V$ is nominally a five-dimensional function of $(s,i,\psi_s,\psi_i,\kappa)$, we are only interested in the Nash solution, for which $\psi=\theta$, which reduces the degrees of freedom by two. Furthermore, to evaluate the $\kappa$ dependance, and the degree to which our neural network can extrapolate beyond the training data, we have evaluated the network on the $\theta$ values given by the pandemic / training trajectory, i.e., only $\kappa$ is allowed to vary. Thus, every point in time along the trajectory, corresponding to a given value $\theta(t)=\theta_t = (s_t, i_t)$, provides a distinct prediction for $V$ as a function of $\kappa$. To visualize this, we color different predictions according to time, or likewise to $s_t$, since the fraction of susceptibles is a monotonically decreasing function of time. The predictions for the $\kappa$ dependence  of the payoff function $V$ are given in Fig.~\eqref{f:l0_uk} for all three values of $\alpha_0$ we have considered. For the smallest value of $\alpha_0$, there is not much information we can obtain from observing the behavior, since there is only a very weak behavior modification $3.5\lesssim \kappa \lesssim \kappa^\star=4$ (i.e., the pandemic proceeds essentially unhindered by behavioural modifications). Even in this extreme case, we obtain relatively good agreement with the exact solution. We can match $V$ on the training points, recover the negative curvature, and obtain the correct order of magnitude. We would like to stress the fact that this payoff function was learned in an unsupervised manner, since $V$ was never included in the training set. Results are even more impressive for higher values of $\alpha_0$, where stronger behaviour modification is observed. In such cases, we are able to recover the correct functional form of the payoff function, this includes the quadratic dependence in $\kappa$, the maximum around $\kappa^\star$, and the pre-factor $\beta = 1$. Furthermore, extrapolation into regions not in the training set, something that could not be expected a priori, is noteworthy. This is likely due to the strong constraints imposed by the optimality condition, and the fact that it must be self-consistently evaluated, which in turn constrain the structure of the neural network encoding the payoff function within the $N^3$.

\subsection{Learning from Behaviour and Dynamics}
\begin{figure}[ht!]
	\centering
	\includegraphics[width=\columnwidth]{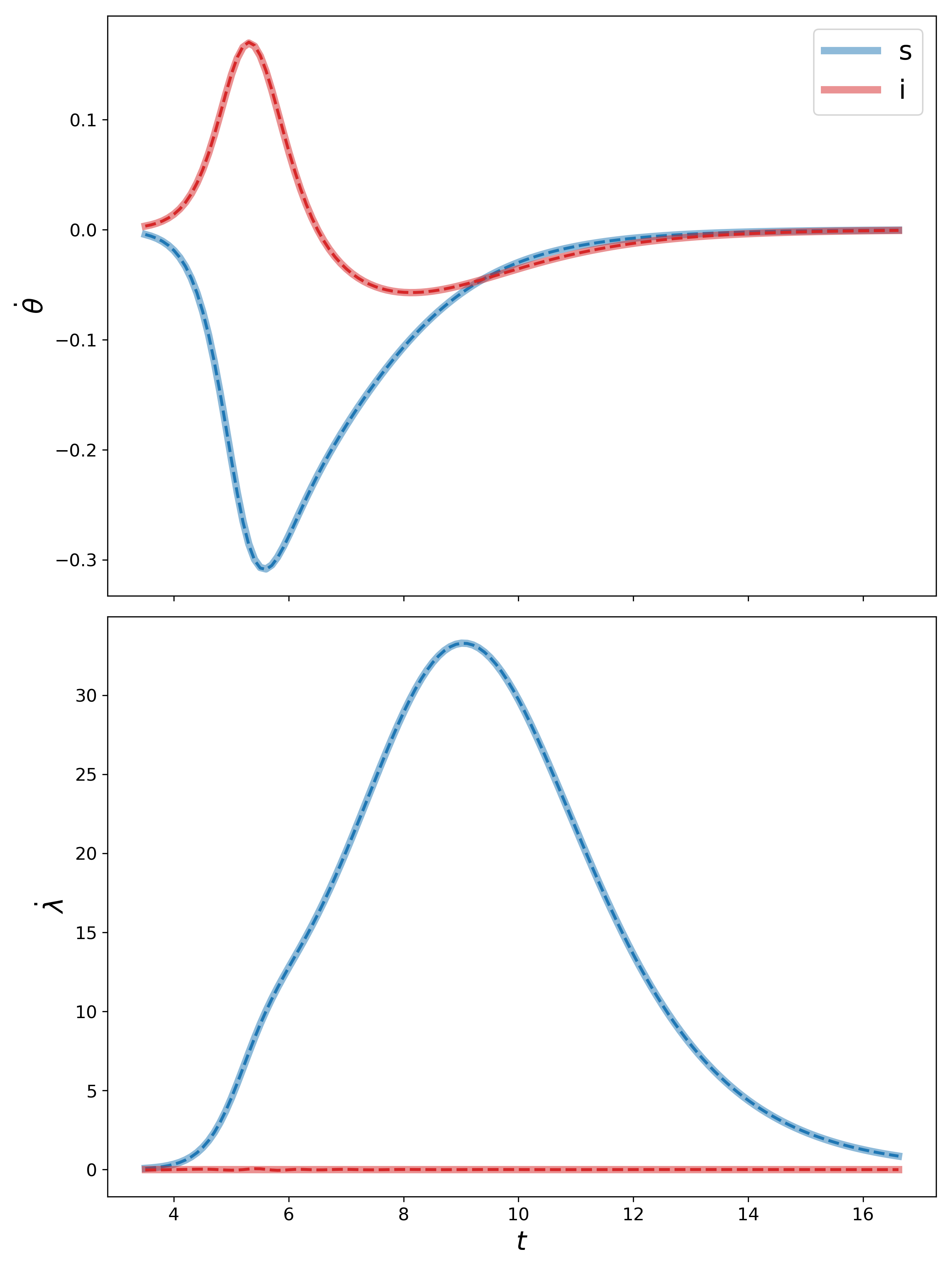}
	\caption{\label{f:d1_l3_dstate} Comparison between the exact (solid) solution and the (dashed) $N^3$ predictions, trained using $\op{L}_{\kappa,\lambda}$, for the state dynamics, $\dot{\theta}$ and $\dot{\lambda}$. Corresponds to the same data as in Fig.~\ref{f:d1_l0_dstate}.}
\end{figure}
\begin{figure}[ht!]
	\centering
	\includegraphics[width=\columnwidth]{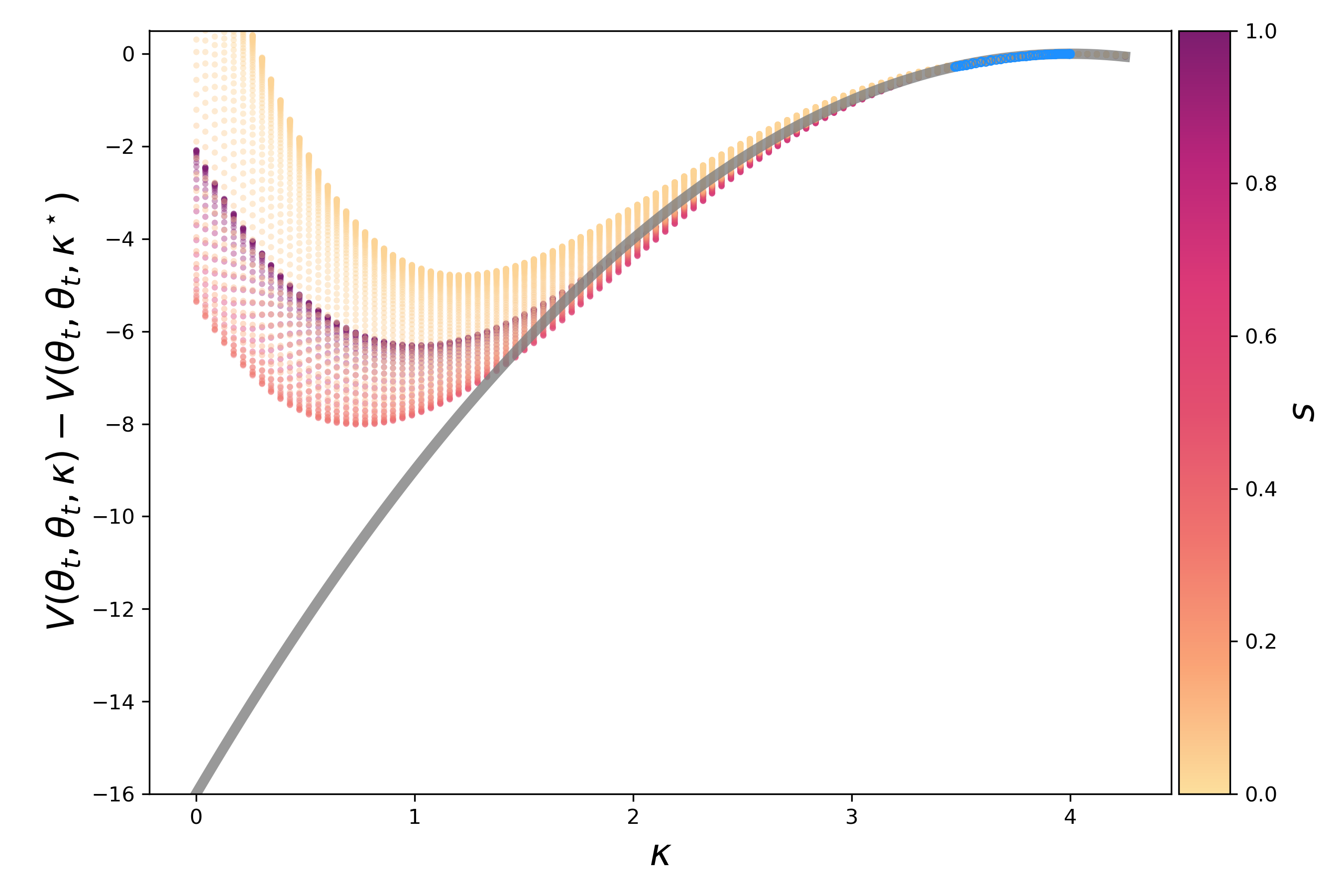}
	\includegraphics[width=\columnwidth]{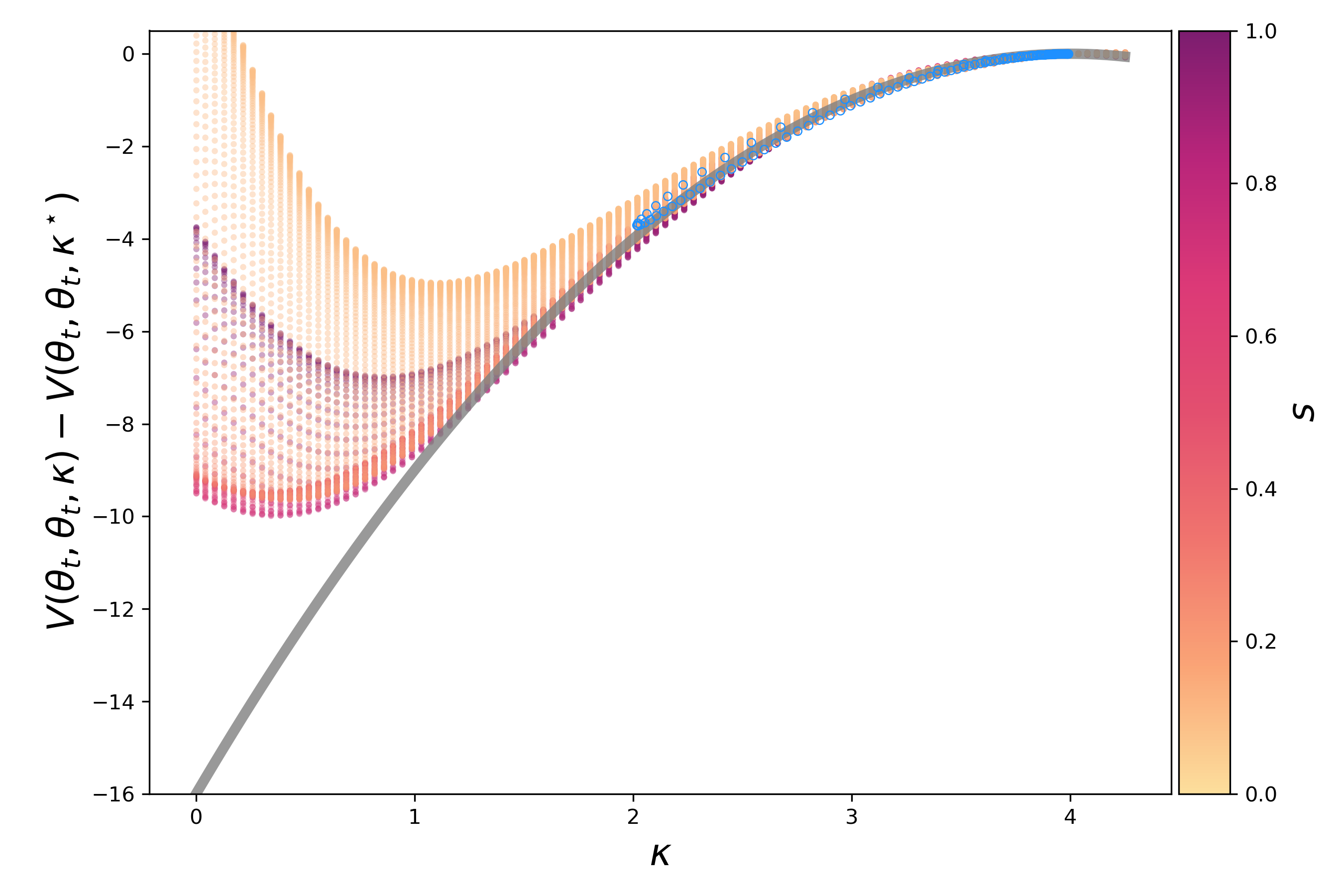}
	\includegraphics[width=\columnwidth]{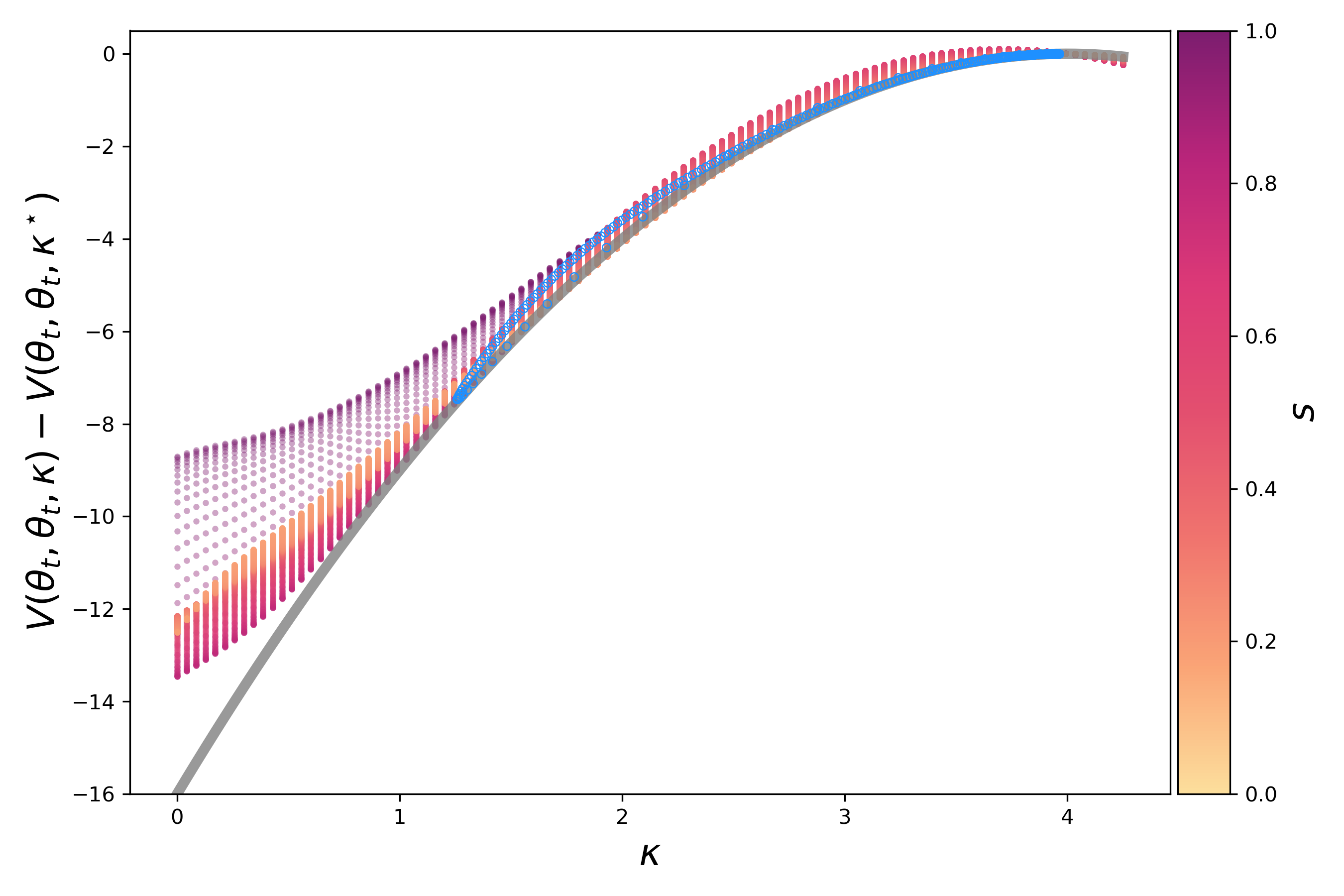}
	\caption{\label{f:l3_uk} $N^3$ predictions for the payoff $V$, as a function of $\kappa$, evaluated at distinct points along the pandemic trajectory, for the three values of $\alpha$, (top) $100$, (middle) $200$, and (bottom) $400$. Similar to Fig.~\ref{f:l0_uk}, but the network was trained using the $\op{L}_{\kappa,\lambda}$ loss function.}
\end{figure}

We now consider the results obtained by learning against both the optimal behaviour $\kappa_\opt$ and the $\lambda$ dynamics, using $\op{L}_{\kappa,\lambda}$.
We obtain excellent agreement for both $\dot{\theta}$ and $\dot{\lambda}$, as seen in Fig.~\ref{f:d1_l3_dstate}. In particular, we are now able to recover the correct dynamics for $\lambda_i$, something that was impossible when only training on $\kappa$. The dynamics is not fitted directly, but through the unknown payoff function, in such a way that the dynamical constraints and Nash equilibrium conditions are satisfied. We obtain similar level of agreement for the $\kappa$ dependence of the payoff $V$, as shown in Fig.~\ref{f:l3_uk}. While at first glance it seems as if the results are not as good as those obtained with $\op{L}_\kappa$, especially at high $\alpha_0$, this is due to the fact that we have fixed the number of training steps. Having to account for both the optimal behaviour and the dynamics results in a more complex learning task, which would require that we train longer to achieve the same level of accuracy.

\begin{figure}[ht!]
	\centering
	\includegraphics[width=\columnwidth]{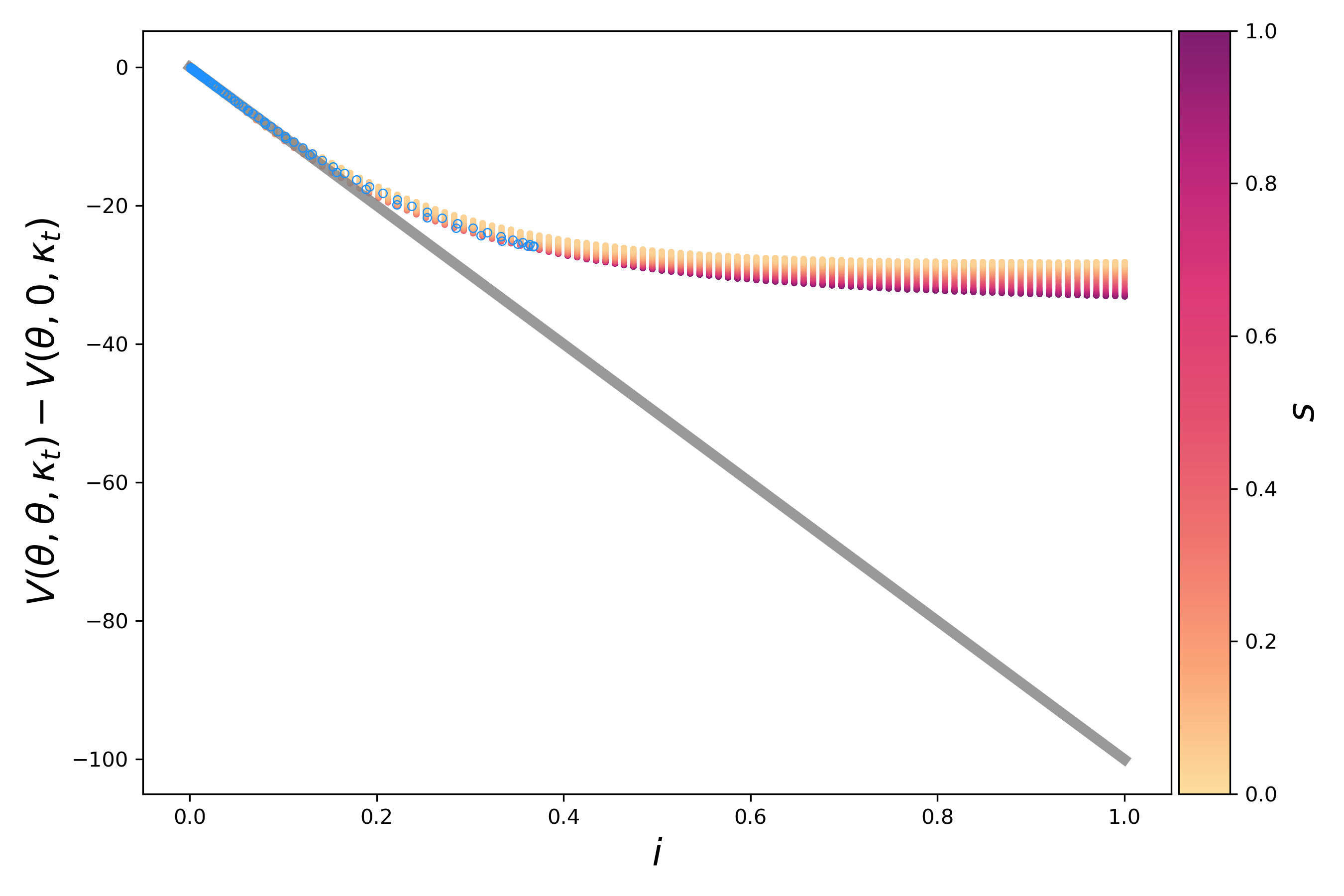}
	\includegraphics[width=\columnwidth]{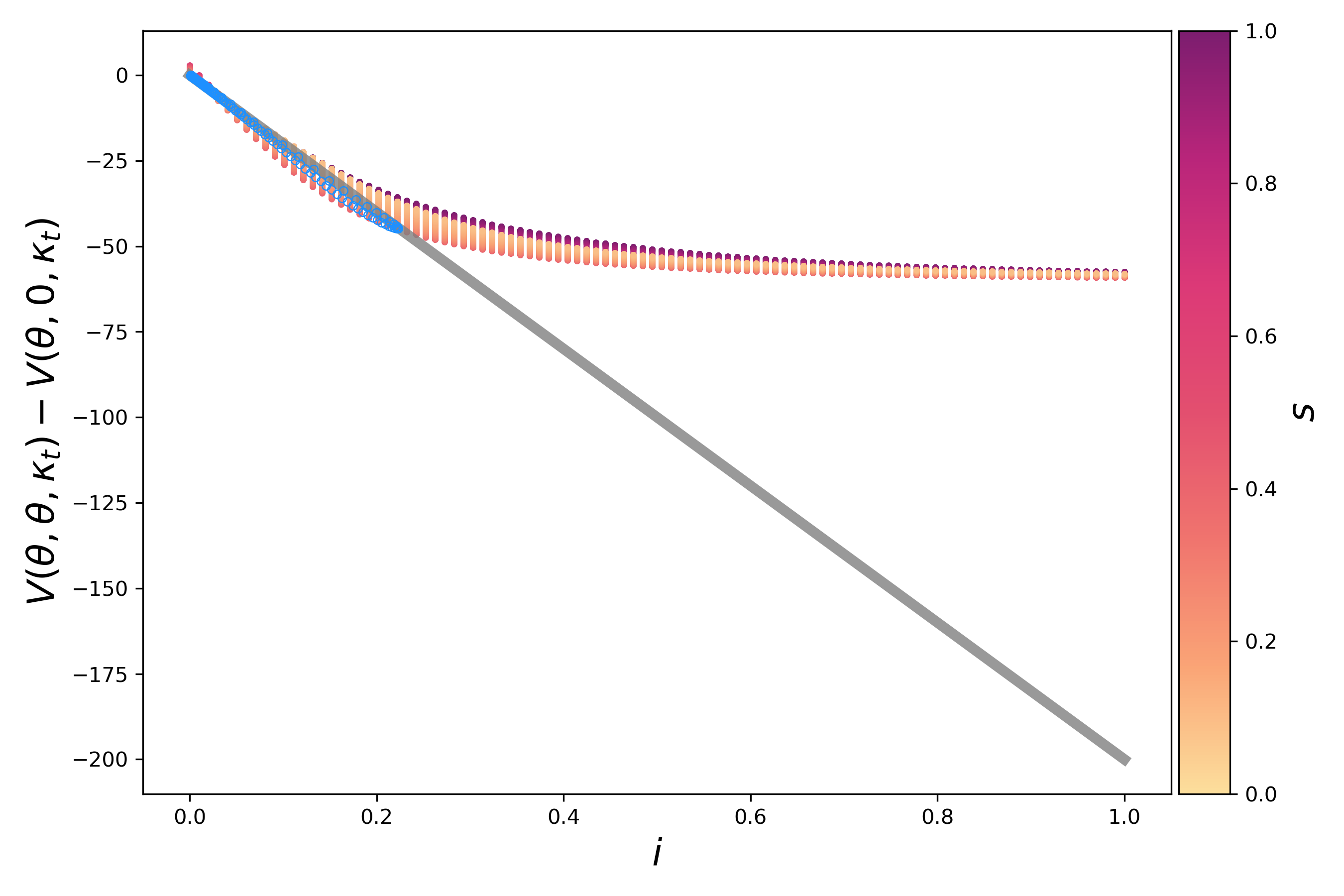}
	\includegraphics[width=\columnwidth]{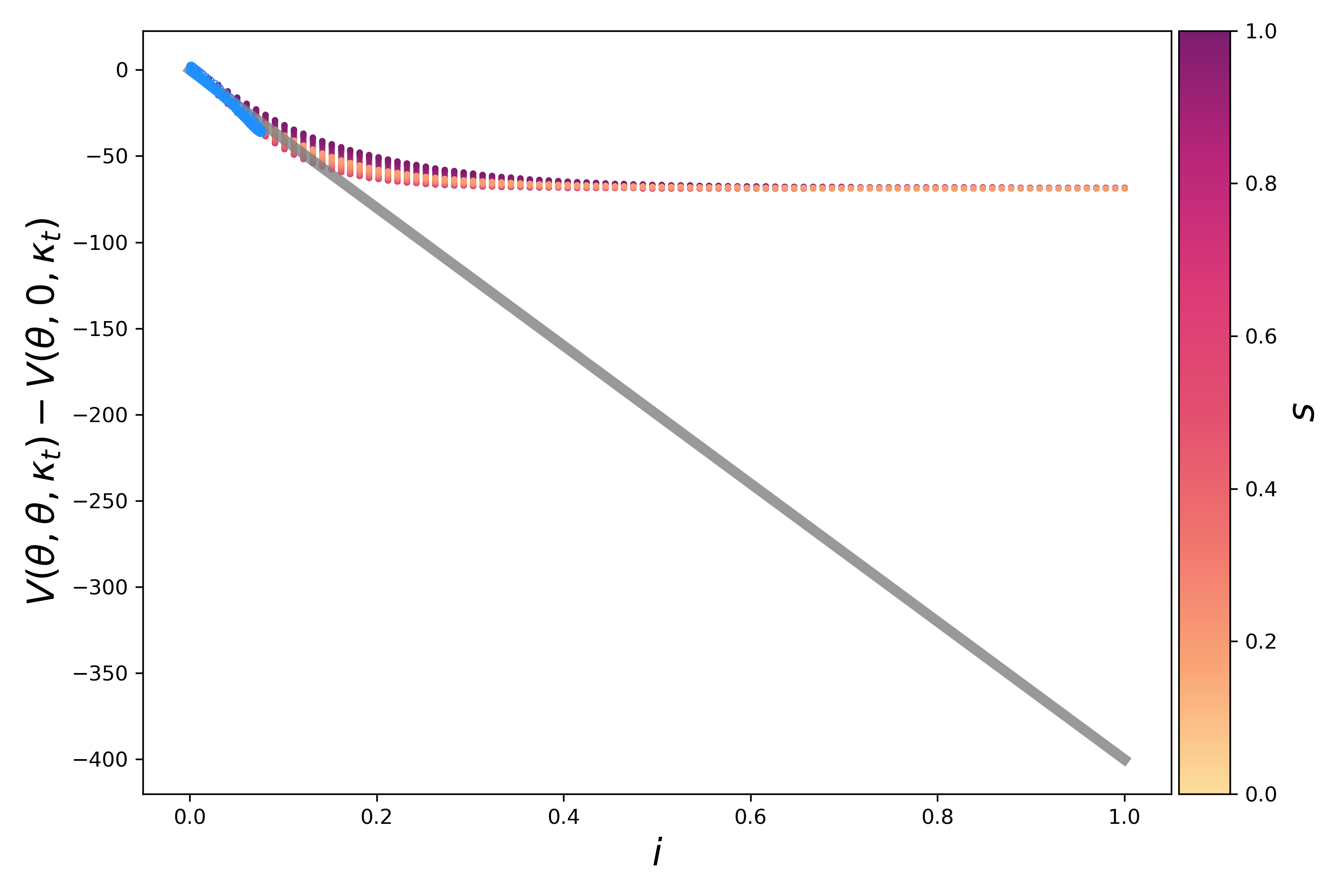}
	\caption{\label{f:l3_ui} $N^3$ predictions for the payoff $V$ as a function of $i$, evaluated at distinct points along the pandemic trajectory, for the three values of $\alpha$, (top) $100$, (middle) $200$, and (bottom) $400$. Similar to Fig.~\ref{f:l0_uk}~and~\ref{f:l3_uk}, but the shift function is modified to suppress the $\kappa$ dependence. The theoretical value (solid line) is now given by $-\alpha \psi_i$.}
\end{figure}

Having access to the full set of data during the training (i.e., $\theta$, $\psi$, $\kappa$ and $\lambda$) means that we are now in a position to recover the full functional dependence of the payoff function. This is illustrated in Fig.~\ref{f:l3_ui}, which shows the shifted potential as a function of the fraction of infected. Not only do we recover the linear behaviour in $\psi_i$, but we also correctly recover the slope, equal to $-\alpha_0$, at least in regions where we have training data. Without further assumptions on the form the payoff function, as provided for the $\kappa$ dependence by the optimality condition, we cannot expect to do any better in predicting the $\psi_i$ ($\psi_s$) dependence, at least for this particular type of single-shot training on individual pandemic trajectories.
\section{Discussion \& Conclusions}
We have developed a new class of neural network, which we have called a Nash Neural Network ($N^3$), that can be applied to optimal control problems in order to infer the underlying payoff function that determines the optimal behaviour. The network is constructed in such a way that (1) it respects the dynamical constraints of the system, and (2) evaluates the payoff at the self-consistently determined optimal behaviour. This is particularly useful when considering differential games with Nash equilibrium. To test our method, we have considered the problem of social distancing during epidemics. The course of the epidemic is assumed to be given by an SIR model that determines the state of the population in terms of the fraction of infected and susceptible. Against this backdrop, we consider rational individuals, whose behaviour is determined by self-interest, so as to maximize their total utility (obtained as the integral of a payoff function), which in our case includes the cost of becoming infected and the cost of socially distancing. The solution to this problem, i.e., what is the optimal behaviour, is well known and has been studied elsewhere\cite{Reluga2010,Schnyder2022}. Here, we considered the inverse problem, that of inferring the utility from the time-evolution of the trajectory. Without making any assumptions regarding the functional form of the payoff, we were able to recover the original function with remarkable accuracy. In particular, we showed that training against the optimal behaviour only, we could recover the terms in the payoff function that included this optimal behaviour (i.e., the social distancing cost). We were even able to extrapolate into regions that were far from the training data. By training against the full set of dynamical variables, we were also able to recover the functional dependence of the payoff function on the state of the system (here the fraction of susceptible and infected).

As currently formulated, the $N^3$ requires as input the full set of dynamical variables, including the Lagrange multipliers $\lambda$ used to constrain the dynamics. Unfortunately, this information will never be available in real-life. However, by considering the boundary values of $\lambda$ as additional hyper-parameters, and integrating our neural network, we can remove the $\lambda$ dependence entirely, and train only on the observed pandemic trajectory. This will be considered in future work, where we will also study the role of noisy measurements in the predictions, as well as introduce the government as an additional player, and simultaneously learn both government and individual utilities. We believe the current work has great potential for many applications in the social and physical sciences, engineering and government, as examples of differential games are ubiquitious\cite{Aumann1992,Myerson1997,Dockner2000,Lambertini2018,McNamara2020,Bauso2016,Nisan2007}.
\acknowledgments
The authors would like to thank T. Taniguchi and A. Oswald for fruitful discussions. This work was supported by the Japan Society for the Promotion of Science (Grants-in-Aid for Scientific Research KAKENHI No. 17K17825 and 20K03786), the SPIRITS 2020 of Kyoto University, and the ``Joint Usage/Research Center for Interdisciplinary Large-scale Information Infrastructures'' and ``High Performance Computing Infrastructure'' in Japan (Project ID:jh210017-MDH). The simulation and machine-learning code was written in Python/JAX\cite{Bradbury2018}, using the JAXopt optimizer library\cite{Blondel2021} to propagate the derivatives through the optimization procedure required to compute the optimal behaviour $\kappa_\opt$. Figures were generated using the Matplotlib\cite{Hunter2007} Python library.
\appendix
\section{Hamiltonian Formulation}\label{s:hamilton}
We briefly discuss the Hamiltonian formulation of the optimal control problem (commonly encountered in the economics and optimal control literature), which we have presented in the main text within the Lagrangian formulation, and show how, in this case, both lead to exactly the same set of equations. The augmented Lagrangian $L(q,\dot{q})$ is a function of generalized coordinates $q=(q_\psi,q_\kappa, q_\lambda)=(\psi,\kappa,\lambda)$ and velocities $\dot{q}=(\dot{q}_\psi, \dot{q}_\kappa, \dot{q}_\lambda)$, and the Hamiltonian $H(q,p)$, obtained from the Legendre transformation $H=-L + \dot{q}\cdot p$\cite{sicm}, is a function of the coordinates $q$ and their conjugate momenta $p$. We have dropped the time $t$ and population state coordinates $\theta$, since they are passive arguments in the Legendre transformation. The momenta are defined as\cite{sicm}
\begin{align}
	p =\begin{pmatrix}
		   p_\psi \\ p_\kappa \\ p_\lambda
	   \end{pmatrix} & \equiv \partial_{\dot{q}}L  = \begin{pmatrix}-q_\lambda\\0\\0\end{pmatrix}
\end{align}
and Hamilton's equations of motion are
\begin{align}
	\dot{q} & = \partial_p H   \\
	\dot{p} & = -\partial_q H.
\end{align}
Because of the particular velocity dependence of the Lagrangian $L$, it turns out that the Hamiltonian is a function of only $q_{\exclude{\lambda}} = (q_\psi,q_\kappa)$ and $p_{\exclude{\lambda}} = (p_\psi, p_\lambda)$, where we use $\exclude{\lambda}$ to indicate that the $\lambda$ coordinates are missing. Expressed as a function of $q$ and $p$, with $\dot{Q}(q,p)$ the function that locally inverts coordinates and momenta to yield the velocities $\dot{q}$,
\begin{align}
	H(q,p) & = -L(q,\dot{Q}(q,p)) + \dot{Q}(q,p)\cdot p                                                                                   \\
	       & = -V(q_{\exclude{\lambda}}) - q_\lambda\cdot \left(F(q_{\exclude{\lambda}}) - \dot{Q}_\psi(q,p)\right)                       \\
	       & \qquad + \dot{Q}_\psi(q,p)\cdot p_\psi \notag                                                                                \\
	       & = -\big(V(q_{\exclude{\lambda}}) + q_\lambda\cdot F(q_{\exclude{\lambda}})\big) + \dot{Q}_\psi(q,p)\cdot(p_\psi + q_\lambda) \\
	       & = -\left(V(q_{\exclude{\lambda}}) - p_\psi \cdot F(q_{\exclude{\lambda}})\right)                                             \\
	       & \equiv H(q_{\exclude{\lambda}},p_{\exclude{\lambda}})
\end{align}
where, in the second to last step we have used the definition of the momenta conjugate to $\psi$, $p_\psi = -\lambda$. From this, we clearly see that, expressed in a Hamiltonian formalism, the Lagrange multipliers $\lambda$ no longer appear as coordinates, but as conjugate momenta. As mentioned in the main text, this Hamiltonian is equivalent to the $L^\prime$ function from which all the dynamical equations were eventually derived
\begin{align}
	H(q_{\exclude{\lambda}},p_{\exclude{\lambda}}) & = -\left(V(q_{\exclude{\lambda}}) - p_\psi \cdot F(q_{\exclude{\lambda}})\right) \\
	                                               & = -L^\prime(q_\psi, q_\kappa, -p_\psi).
\end{align}
Finally, Hamilton's equations for the $\psi$ degrees of freedom provide both the constraint and $\lambda$ dynamics,
\begin{align}
	\dot{\psi}                      & = F (\psi, \kappa, \lambda)                       \\
	\dot{p}_{\psi} = -\dot{\lambda} & = \partial_{\psi} L^\prime(\psi, \kappa, \lambda)
\end{align}
whereas the optimality condition is derived from the momentum equation for the remaining $\kappa$ degree of freedom,
\begin{align}
	p_\kappa                       & = 0                                                     \\
	\Longrightarrow \dot{p}_\kappa & = \partial_\kappa L^\prime(\psi, \kappa, \lambda)  = 0.
\end{align}
\section{Non-Uniqueness of the Lagrangian / Payoff Function}\label{s:uniqueness}
It is well known that Lagrangians are not unique\cite{Goldstein2001,Arnold1989,sicm}, which raises the question of how we can expect to recover the payoff function $V$ from observations of the dynamics. In particular, two Lagrangians $L$ and $\alternate{L}$ that differ in the total-time derivative of a function of time and coordinates, $\alternate{L}-L = D_t G(t,q)$, will give rise to the same Euler-Lagrange equations\cite{sicm}. This is easily seen, as both Lagrangians lead to the same action integral, except for a difference in the end-point values
\begin{align}
	\alternate{S}[q](t_1,t_2) & = \int_{t_1}^{t_2} \left\{L(t,q,\dot{q})+ D_t G(t,q)\right\} \dd{t} \\
	                          & = S[q](t_1,t_2) + \left.G(t,q)\right\rvert_{t_1}^{t_2}.
\end{align}
Variations which extremize $S$ would also extremize $\alternate{S}$,
\begin{align}
	\delta_q\alternate{S}[q](t_1,t_2) & = \delta_q S[q](t_1,t_2) + \left.\left(\partial_q G\right)\delta \eta\right\rvert_{t_1}^{t_2}
\end{align}
only changing the natural boundary conditions. However, for the particular problem we are considering, with Lagrangians of the form
\begin{align}
	L(t, q, \dot{q}) & = V(t, q_{\exclude{\lambda}}) + q_\lambda\cdot\left(F(t,q_{\exclude{\lambda}}) - \dot{q}_\psi\right)
\end{align}
it is impossible to accommodate a general total-time derivative, since
\begin{align}
	D_t G(t, q) & = \partial_t G(t,q) + \dot{q}\cdot\partial_q G(t,q)                                                   \\
	            & = \partial_t G(t,q) + \dot{q}_\psi\cdot\partial_{\psi} G(t, q)                                        \\
	            & \quad+\dot{q}_\kappa\cdot\partial_\kappa G(t,q) + \dot{q}_\lambda\cdot \partial_\lambda G(t,q).\notag
\end{align}
By construction, the terms linear in $\dot{q}$ are not allowed, since neither $V$ nor $F$ depend on $\dot{q}$, and the dependence on $\dot{q}_\psi$ in $L$ is exactly given by the constraint term $\lambda\cdot(F-\dot{q}_\psi)$. Thus, the only possibility would be to consider a function that is independent of $q$. The general form of the Lagrangian / payoff function is then
\begin{align}
	\alternate{L}(t; \theta, \psi, \kappa, \lambda; \cdot, \dot{\psi}, \cdot, \cdot) & = \alternate{V}(t, \theta, \psi, \kappa)                                 \\
	                                                                                 & + \lambda\cdot\left(F(t, \theta, \psi, \kappa)- \dot{\psi}\right) \notag \\
	\alternate{V}(t, \theta, \psi, \kappa)                                           & = V(t, \theta, \psi, \kappa) + G^\prime(t,\theta),
\end{align}
with $G^\prime(t,\theta) = D_t G(t,\theta) = \partial_t G(t, \theta)$ an arbitrary function of time and population state variables $\theta$ only. In conclusion, since $G(t,\theta)$  is independent of the individual state variables $q=(\psi,\kappa,\lambda)$, it has absolutely no effect on the derived dynamics, and cannot be recovered from such observations. Thus, when evaluating the neural network predictions for $V$, it only makes sense to look at changes in $V$, in order to remove the $G(t,\theta)$ term that effectively defines the zero of this ``potential energy''.

\begin{thebibliography}{84}
	\expandafter\ifx\csname natexlab\endcsname\relax\def\natexlab#1{#1}\fi
	\expandafter\ifx\csname bibnamefont\endcsname\relax
		\def\bibnamefont#1{#1}\fi
	\expandafter\ifx\csname bibfnamefont\endcsname\relax
		\def\bibfnamefont#1{#1}\fi
	\expandafter\ifx\csname citenamefont\endcsname\relax
		\def\citenamefont#1{#1}\fi
	\expandafter\ifx\csname url\endcsname\relax
		\def\url#1{\texttt{#1}}\fi
	\expandafter\ifx\csname urlprefix\endcsname\relax\def\urlprefix{URL }\fi
	\providecommand{\bibinfo}[2]{#2}
	\providecommand{\eprint}[2][]{\url{#2}}

	\bibitem[{\citenamefont{Aumann and Hart}(1992)}]{Aumann1992}
	\bibinfo{editor}{\bibfnamefont{R.~J.} \bibnamefont{Aumann}} \bibnamefont{and}
	\bibinfo{editor}{\bibfnamefont{S.}~\bibnamefont{Hart}}, eds.,
	\emph{\bibinfo{title}{{Handbook of Game Theory with Economic Applications}}}
	(\bibinfo{publisher}{North-Holland}, \bibinfo{address}{Amsterdam},
	\bibinfo{year}{1992}).

	\bibitem[{\citenamefont{Myerson}(1997)}]{Myerson1997}
	\bibinfo{author}{\bibfnamefont{R.~B.} \bibnamefont{Myerson}},
	\emph{\bibinfo{title}{{Game Theory : Analysis of Conflict}}}
	(\bibinfo{publisher}{Harvard University Press}, \bibinfo{address}{Cambridge},
	\bibinfo{year}{1997}).

	\bibitem[{\citenamefont{Dockner et~al.}(2000)\citenamefont{Dockner, Jorgensen,
					{Van Long}, and Sorger}}]{Dockner2000}
	\bibinfo{author}{\bibfnamefont{E.~J.} \bibnamefont{Dockner}},
	\bibinfo{author}{\bibfnamefont{S.}~\bibnamefont{Jorgensen}},
	\bibinfo{author}{\bibfnamefont{N.}~\bibnamefont{{Van Long}}},
	\bibnamefont{and} \bibinfo{author}{\bibfnamefont{G.}~\bibnamefont{Sorger}},
	\emph{\bibinfo{title}{{Differential Games in Economics and Management
					Science}}} (\bibinfo{publisher}{Cambridge University Press},
	\bibinfo{address}{Cambridge}, \bibinfo{year}{2000}).

	\bibitem[{\citenamefont{Lambertini}(2018)}]{Lambertini2018}
	\bibinfo{author}{\bibfnamefont{L.}~\bibnamefont{Lambertini}},
	\emph{\bibinfo{title}{{Differential Games in Industrial Economics}}}
	(\bibinfo{publisher}{Cambridge University Press},
	\bibinfo{address}{Cambridge}, \bibinfo{year}{2018}), \bibinfo{edition}{1st}
	ed.

	\bibitem[{\citenamefont{McNamara and Leima}(2020)}]{McNamara2020}
	\bibinfo{author}{\bibfnamefont{J.~M.} \bibnamefont{McNamara}} \bibnamefont{and}
	\bibinfo{author}{\bibfnamefont{O.}~\bibnamefont{Leima}},
	\emph{\bibinfo{title}{{Game Theory in Biology}}} (\bibinfo{publisher}{Oxford
		University Press}, \bibinfo{address}{Oxford}, \bibinfo{year}{2020}).

	\bibitem[{\citenamefont{Bauso}(2016)}]{Bauso2016}
	\bibinfo{author}{\bibfnamefont{D.}~\bibnamefont{Bauso}},
	\emph{\bibinfo{title}{{Game Theory with Engineering Applications}}}
	(\bibinfo{publisher}{Society for Industrial and Applied Mathematics},
	\bibinfo{address}{Philadelphia}, \bibinfo{year}{2016}).

	\bibitem[{\citenamefont{Nisan et~al.}(2007)\citenamefont{Nisan, Roughgarden,
					Tardos, and Vazirani}}]{Nisan2007}
	\bibinfo{editor}{\bibfnamefont{N.}~\bibnamefont{Nisan}},
	\bibinfo{editor}{\bibfnamefont{T.}~\bibnamefont{Roughgarden}},
	\bibinfo{editor}{\bibfnamefont{E.}~\bibnamefont{Tardos}}, \bibnamefont{and}
	\bibinfo{editor}{\bibfnamefont{V.~V.} \bibnamefont{Vazirani}}, eds.,
	\emph{\bibinfo{title}{{Algorithmic Game Theory}}}
	(\bibinfo{publisher}{Cambridge University Press}, \bibinfo{address}{New
		York}, \bibinfo{year}{2007}).

	\bibitem[{\citenamefont{Brunton and Kutz}(2019)}]{Brunton2019}
	\bibinfo{author}{\bibfnamefont{S.~L.} \bibnamefont{Brunton}} \bibnamefont{and}
	\bibinfo{author}{\bibfnamefont{J.~N.} \bibnamefont{Kutz}},
	\emph{\bibinfo{title}{{Data-Driven Science and Engineering}}}
	(\bibinfo{publisher}{Cambridge University Press},
	\bibinfo{address}{Cambridge}, \bibinfo{year}{2019}), \bibinfo{edition}{1st}
	ed.

	\bibitem[{\citenamefont{Raissi et~al.}(2019)\citenamefont{Raissi, Perdikaris,
					and Karniadakis}}]{Raissi2019a}
	\bibinfo{author}{\bibfnamefont{M.}~\bibnamefont{Raissi}},
	\bibinfo{author}{\bibfnamefont{P.}~\bibnamefont{Perdikaris}},
	\bibnamefont{and} \bibinfo{author}{\bibfnamefont{G.~E.}
		\bibnamefont{Karniadakis}}, \bibinfo{journal}{Journal of Computational
		Physics} \textbf{\bibinfo{volume}{378}}, \bibinfo{pages}{686}
	(\bibinfo{year}{2019}).

	\bibitem[{\citenamefont{Karniadakis et~al.}(2021)\citenamefont{Karniadakis,
					Kevrekidis, Lu, Perdikaris, Wang, and Yang}}]{Karniadakis2021}
	\bibinfo{author}{\bibfnamefont{G.~E.} \bibnamefont{Karniadakis}},
	\bibinfo{author}{\bibfnamefont{I.~G.} \bibnamefont{Kevrekidis}},
	\bibinfo{author}{\bibfnamefont{L.}~\bibnamefont{Lu}},
	\bibinfo{author}{\bibfnamefont{P.}~\bibnamefont{Perdikaris}},
	\bibinfo{author}{\bibfnamefont{S.}~\bibnamefont{Wang}}, \bibnamefont{and}
	\bibinfo{author}{\bibfnamefont{L.}~\bibnamefont{Yang}},
	\bibinfo{journal}{Nature Reviews Physics} \textbf{\bibinfo{volume}{3}},
	\bibinfo{pages}{422} (\bibinfo{year}{2021}).

	\bibitem[{\citenamefont{Greydanus et~al.}(2019)\citenamefont{Greydanus, Dzamba,
					and Yosinski}}]{Greydanus2019}
	\bibinfo{author}{\bibfnamefont{S.}~\bibnamefont{Greydanus}},
	\bibinfo{author}{\bibfnamefont{M.}~\bibnamefont{Dzamba}}, \bibnamefont{and}
	\bibinfo{author}{\bibfnamefont{J.}~\bibnamefont{Yosinski}},
	\bibinfo{journal}{Advances in Neural Information Processing Systems}
	\textbf{\bibinfo{volume}{32}}, \bibinfo{pages}{1} (\bibinfo{year}{2019}),
	\eprint{arXiv:1906.01563}.

	\bibitem[{\citenamefont{Bertalan et~al.}(2019)\citenamefont{Bertalan, Dietrich,
	Mezi{\'{c}}, and Kevrekidis}}]{Bertalan2019}
	\bibinfo{author}{\bibfnamefont{T.}~\bibnamefont{Bertalan}},
	\bibinfo{author}{\bibfnamefont{F.}~\bibnamefont{Dietrich}},
	\bibinfo{author}{\bibfnamefont{I.}~\bibnamefont{Mezi{\'{c}}}},
	\bibnamefont{and} \bibinfo{author}{\bibfnamefont{I.~G.}
		\bibnamefont{Kevrekidis}}, \bibinfo{journal}{Chaos}
	\textbf{\bibinfo{volume}{29}} (\bibinfo{year}{2019}), \eprint{1907.12715}.

	\bibitem[{\citenamefont{Zhong et~al.}(2020{\natexlab{a}})\citenamefont{Zhong,
					Dey, and Chakraborty}}]{Zhong2019}
	\bibinfo{author}{\bibfnamefont{Y.~D.} \bibnamefont{Zhong}},
	\bibinfo{author}{\bibfnamefont{B.}~\bibnamefont{Dey}}, \bibnamefont{and}
	\bibinfo{author}{\bibfnamefont{A.}~\bibnamefont{Chakraborty}}, in
	\emph{\bibinfo{booktitle}{International Conference on Learning
			Representations (ICLR)}} (\bibinfo{year}{2020}{\natexlab{a}}),
	\eprint{arXiv:1909.12077}, \urlprefix\url{http://arxiv.org/abs/1909.12077}.

	\bibitem[{\citenamefont{Cranmer et~al.}(2020)\citenamefont{Cranmer, Greydanus,
					Hoyer, Battaglia, Spergel, and Ho}}]{Cranmer2020}
	\bibinfo{author}{\bibfnamefont{M.}~\bibnamefont{Cranmer}},
	\bibinfo{author}{\bibfnamefont{S.}~\bibnamefont{Greydanus}},
	\bibinfo{author}{\bibfnamefont{S.}~\bibnamefont{Hoyer}},
	\bibinfo{author}{\bibfnamefont{P.}~\bibnamefont{Battaglia}},
	\bibinfo{author}{\bibfnamefont{D.}~\bibnamefont{Spergel}}, \bibnamefont{and}
	\bibinfo{author}{\bibfnamefont{S.}~\bibnamefont{Ho}}, \bibinfo{journal}{arXiv
		preprint}  (\bibinfo{year}{2020}), \eprint{arXiv:2003.04630},
	\urlprefix\url{http://arxiv.org/abs/2003.04630}.

	\bibitem[{\citenamefont{Zhong et~al.}(2020{\natexlab{b}})\citenamefont{Zhong,
					Dey, and Chakraborty}}]{Zhong2020}
	\bibinfo{author}{\bibfnamefont{Y.~D.} \bibnamefont{Zhong}},
	\bibinfo{author}{\bibfnamefont{B.}~\bibnamefont{Dey}}, \bibnamefont{and}
	\bibinfo{author}{\bibfnamefont{A.}~\bibnamefont{Chakraborty}},
	\bibinfo{journal}{International Conference on Learning Representations
		(ICLR)}  (\bibinfo{year}{2020}{\natexlab{b}}), \eprint{arXiv:2002.08860},
	\urlprefix\url{http://arxiv.org/abs/2002.08860}.

	\bibitem[{\citenamefont{Choudhary et~al.}(2020)\citenamefont{Choudhary,
					Lindner, Holliday, Miller, Sinha, and Ditto}}]{Choudhary2020}
	\bibinfo{author}{\bibfnamefont{A.}~\bibnamefont{Choudhary}},
	\bibinfo{author}{\bibfnamefont{J.~F.} \bibnamefont{Lindner}},
	\bibinfo{author}{\bibfnamefont{E.~G.} \bibnamefont{Holliday}},
	\bibinfo{author}{\bibfnamefont{S.~T.} \bibnamefont{Miller}},
	\bibinfo{author}{\bibfnamefont{S.}~\bibnamefont{Sinha}}, \bibnamefont{and}
	\bibinfo{author}{\bibfnamefont{W.~L.} \bibnamefont{Ditto}},
	\bibinfo{journal}{Physical Review E} \textbf{\bibinfo{volume}{101}},
	\bibinfo{pages}{1} (\bibinfo{year}{2020}).

	\bibitem[{\citenamefont{Lee and Seong}(2020)}]{Lee2020}
	\bibinfo{author}{\bibfnamefont{S.}~\bibnamefont{Lee}} \bibnamefont{and}
	\bibinfo{author}{\bibfnamefont{W.}~\bibnamefont{Seong}},
	\bibinfo{journal}{34th Conference on Neural Information Processing Systems
		(NeurIPS)}  (\bibinfo{year}{2020}).

	\bibitem[{\citenamefont{Finzi et~al.}(2020)\citenamefont{Finzi, Wang, and
					Wilson}}]{Finzi2020}
	\bibinfo{author}{\bibfnamefont{M.}~\bibnamefont{Finzi}},
	\bibinfo{author}{\bibfnamefont{K.~A.} \bibnamefont{Wang}}, \bibnamefont{and}
	\bibinfo{author}{\bibfnamefont{A.~G.} \bibnamefont{Wilson}},
	\bibinfo{journal}{34th Conference on Neural Information Processing Systems
		(NeurIPS)}  (\bibinfo{year}{2020}), \eprint{arXiv:2010.13581}.

	\bibitem[{\citenamefont{Roehrl et~al.}(2020)\citenamefont{Roehrl, Runkler,
					Brandtstetter, Tokic, and Obermayer}}]{Roehrl2020}
	\bibinfo{author}{\bibfnamefont{M.~A.} \bibnamefont{Roehrl}},
	\bibinfo{author}{\bibfnamefont{T.~A.} \bibnamefont{Runkler}},
	\bibinfo{author}{\bibfnamefont{V.}~\bibnamefont{Brandtstetter}},
	\bibinfo{author}{\bibfnamefont{M.}~\bibnamefont{Tokic}}, \bibnamefont{and}
	\bibinfo{author}{\bibfnamefont{S.}~\bibnamefont{Obermayer}},
	\bibinfo{journal}{IFAC-PapersOnLine} \textbf{\bibinfo{volume}{53}},
	\bibinfo{pages}{9195} (\bibinfo{year}{2020}), \eprint{arXiv:2005.14617},
	\urlprefix\url{https://doi.org/10.1016/j.ifacol.2020.12.2182}.

	\bibitem[{\citenamefont{Lutter and Peters}(2021)}]{Lutter2021}
	\bibinfo{author}{\bibfnamefont{M.}~\bibnamefont{Lutter}} \bibnamefont{and}
	\bibinfo{author}{\bibfnamefont{J.}~\bibnamefont{Peters}},
	\bibinfo{journal}{arXiv preprint}  (\bibinfo{year}{2021}),
	\eprint{arXiv:2110.01894}, \urlprefix\url{http://arxiv.org/abs/2110.01894}.

	\bibitem[{\citenamefont{Zhong et~al.}(2021)\citenamefont{Zhong, Dey, and
					Chakraborty}}]{Zhong2021}
	\bibinfo{author}{\bibfnamefont{Y.~D.} \bibnamefont{Zhong}},
	\bibinfo{author}{\bibfnamefont{B.}~\bibnamefont{Dey}}, \bibnamefont{and}
	\bibinfo{author}{\bibfnamefont{A.}~\bibnamefont{Chakraborty}},
	\bibinfo{journal}{35th Conference on Neural Information Processing Systems
		(NeurIPS 2021)}  (\bibinfo{year}{2021}), \eprint{arXiv:2102.06794},
	\urlprefix\url{http://arxiv.org/abs/2102.06794}.

	\bibitem[{\citenamefont{Duong and Atanasov}(2021)}]{Duong2021}
	\bibinfo{author}{\bibfnamefont{T.}~\bibnamefont{Duong}} \bibnamefont{and}
	\bibinfo{author}{\bibfnamefont{N.}~\bibnamefont{Atanasov}},
	\bibinfo{journal}{arXiv preprint}  (\bibinfo{year}{2021}),
	\eprint{arXiv:2106.12782}.

	\bibitem[{\citenamefont{Han et~al.}(2021)\citenamefont{Han, Glaz, Haile, and
					Lai}}]{Han2021}
	\bibinfo{author}{\bibfnamefont{C.~D.} \bibnamefont{Han}},
	\bibinfo{author}{\bibfnamefont{B.}~\bibnamefont{Glaz}},
	\bibinfo{author}{\bibfnamefont{M.}~\bibnamefont{Haile}}, \bibnamefont{and}
	\bibinfo{author}{\bibfnamefont{Y.~C.} \bibnamefont{Lai}},
	\bibinfo{journal}{Physical Review Research} \textbf{\bibinfo{volume}{3}},
	\bibinfo{pages}{1} (\bibinfo{year}{2021}), \eprint{arXiv:2102.13235}.

	\bibitem[{\citenamefont{Sosanya and Greydanus}(2022)}]{Sosanya2022}
	\bibinfo{author}{\bibfnamefont{A.}~\bibnamefont{Sosanya}} \bibnamefont{and}
	\bibinfo{author}{\bibfnamefont{S.}~\bibnamefont{Greydanus}},
	\bibinfo{journal}{arXiv preprint}  (\bibinfo{year}{2022}),
	\eprint{arXiv:2201.10085}, \urlprefix\url{http://arxiv.org/abs/2201.10085}.

	\bibitem[{\citenamefont{Celledoni et~al.}(2022)\citenamefont{Celledoni, Leone,
					Murari, and Owren}}]{Celledoni2022}
	\bibinfo{author}{\bibfnamefont{E.}~\bibnamefont{Celledoni}},
	\bibinfo{author}{\bibfnamefont{A.}~\bibnamefont{Leone}},
	\bibinfo{author}{\bibfnamefont{D.}~\bibnamefont{Murari}}, \bibnamefont{and}
	\bibinfo{author}{\bibfnamefont{B.}~\bibnamefont{Owren}},
	\bibinfo{journal}{arXiv preprint}  (\bibinfo{year}{2022}),
	\eprint{arXiv:2201.13254}, \urlprefix\url{http://arxiv.org/abs/2201.13254}.

	\bibitem[{\citenamefont{Chen et~al.}(2022)\citenamefont{Chen, Feng, Yan, and
					Zha}}]{Chen2022}
	\bibinfo{author}{\bibfnamefont{Z.}~\bibnamefont{Chen}},
	\bibinfo{author}{\bibfnamefont{M.}~\bibnamefont{Feng}},
	\bibinfo{author}{\bibfnamefont{J.}~\bibnamefont{Yan}}, \bibnamefont{and}
	\bibinfo{author}{\bibfnamefont{H.}~\bibnamefont{Zha}},
	\bibinfo{journal}{arXiv preprint}  (\bibinfo{year}{2022}),
	\eprint{arXiv:2203.00128}, \urlprefix\url{http://arxiv.org/abs/2203.00128}.

	\bibitem[{\citenamefont{Jin et~al.}(2020{\natexlab{a}})\citenamefont{Jin,
					Zhang, Zhu, Tang, and Karniadakis}}]{Jin2020}
	\bibinfo{author}{\bibfnamefont{P.}~\bibnamefont{Jin}},
	\bibinfo{author}{\bibfnamefont{Z.}~\bibnamefont{Zhang}},
	\bibinfo{author}{\bibfnamefont{A.}~\bibnamefont{Zhu}},
	\bibinfo{author}{\bibfnamefont{Y.}~\bibnamefont{Tang}}, \bibnamefont{and}
	\bibinfo{author}{\bibfnamefont{G.~E.} \bibnamefont{Karniadakis}},
	\bibinfo{journal}{Neural Networks} \textbf{\bibinfo{volume}{132}},
	\bibinfo{pages}{166} (\bibinfo{year}{2020}{\natexlab{a}}),
	\eprint{arXiv:2001.03750},
	\urlprefix\url{https://doi.org/10.1016/j.neunet.2020.08.017}.

	\bibitem[{\citenamefont{Chen et~al.}(2020)\citenamefont{Chen, Zhang, Arjovsky,
					and Bottou}}]{Chen2019}
	\bibinfo{author}{\bibfnamefont{Z.}~\bibnamefont{Chen}},
	\bibinfo{author}{\bibfnamefont{J.}~\bibnamefont{Zhang}},
	\bibinfo{author}{\bibfnamefont{M.}~\bibnamefont{Arjovsky}}, \bibnamefont{and}
	\bibinfo{author}{\bibfnamefont{L.}~\bibnamefont{Bottou}}, in
	\emph{\bibinfo{booktitle}{International Conference on Learning
			Representations (ICLR)}} (\bibinfo{year}{2020}), \eprint{arXiv:1909.13334},
	\urlprefix\url{http://arxiv.org/abs/1909.13334}.

	\bibitem[{\citenamefont{Meng et~al.}(2022)\citenamefont{Meng, Zhang, Darbon,
					and Karniadakis}}]{Meng2022}
	\bibinfo{author}{\bibfnamefont{T.}~\bibnamefont{Meng}},
	\bibinfo{author}{\bibfnamefont{Z.}~\bibnamefont{Zhang}},
	\bibinfo{author}{\bibfnamefont{J.}~\bibnamefont{Darbon}}, \bibnamefont{and}
	\bibinfo{author}{\bibfnamefont{G.~E.} \bibnamefont{Karniadakis}},
	\bibinfo{journal}{arXiv preprint}  (\bibinfo{year}{2022}),
	\eprint{arXiv:2201.05475}, \urlprefix\url{http://arxiv.org/abs/2201.05475}.

	\bibitem[{\citenamefont{Erichson et~al.}(2021)\citenamefont{Erichson, Azencot,
					Queiruga, Hodgkinson, and Mahoney}}]{Erichson2020}
	\bibinfo{author}{\bibfnamefont{N.~B.} \bibnamefont{Erichson}},
	\bibinfo{author}{\bibfnamefont{O.}~\bibnamefont{Azencot}},
	\bibinfo{author}{\bibfnamefont{A.}~\bibnamefont{Queiruga}},
	\bibinfo{author}{\bibfnamefont{L.}~\bibnamefont{Hodgkinson}},
	\bibnamefont{and} \bibinfo{author}{\bibfnamefont{M.~W.}
		\bibnamefont{Mahoney}}, in \emph{\bibinfo{booktitle}{International Conference
			on Learning Representations (ICLR)}} (\bibinfo{year}{2021}),
	\eprint{arXiv:2006.12070}, \urlprefix\url{http://arxiv.org/abs/2006.12070}.

	\bibitem[{\citenamefont{Jin et~al.}(2022{\natexlab{a}})\citenamefont{Jin,
					Zhang, Kevrekidis, and Karniadakis}}]{Jin2022}
	\bibinfo{author}{\bibfnamefont{P.}~\bibnamefont{Jin}},
	\bibinfo{author}{\bibfnamefont{Z.}~\bibnamefont{Zhang}},
	\bibinfo{author}{\bibfnamefont{I.~G.} \bibnamefont{Kevrekidis}},
	\bibnamefont{and} \bibinfo{author}{\bibfnamefont{G.~E.}
		\bibnamefont{Karniadakis}}, \bibinfo{journal}{IEEE Transactions on Neural
		Networks and Learning Systems} pp. \bibinfo{pages}{1--13}
	(\bibinfo{year}{2022}{\natexlab{a}}), \eprint{arXiv:2012.03133}.

	\bibitem[{\citenamefont{{\v{S}}{\'{i}}pka and Pavelka}(2021)}]{Sipka2021}
	\bibinfo{author}{\bibfnamefont{M.}~\bibnamefont{{\v{S}}{\'{i}}pka}}
	\bibnamefont{and} \bibinfo{author}{\bibfnamefont{M.}~\bibnamefont{Pavelka}},
	\bibinfo{journal}{arXiv preprint}  (\bibinfo{year}{2021}),
	\eprint{arXiv:2109.12659}, \urlprefix\url{http://arxiv.org/abs/2109.12659}.

	\bibitem[{\citenamefont{Zhang et~al.}(2021)\citenamefont{Zhang, Shin, and
					Karniadakis}}]{Zhang2021a}
	\bibinfo{author}{\bibfnamefont{Z.}~\bibnamefont{Zhang}},
	\bibinfo{author}{\bibfnamefont{Y.}~\bibnamefont{Shin}}, \bibnamefont{and}
	\bibinfo{author}{\bibfnamefont{G.~E.} \bibnamefont{Karniadakis}},
	\bibinfo{journal}{arXiv preprint}  (\bibinfo{year}{2021}),
	\eprint{arXiv:2109.00092}, \urlprefix\url{http://arxiv.org/abs/2109.00092}.

	\bibitem[{\citenamefont{Alet et~al.}(2021)\citenamefont{Alet, Doblar, Zhou,
					Tenenbaum, Kawaguchi, and Finn}}]{Alet2021}
	\bibinfo{author}{\bibfnamefont{F.}~\bibnamefont{Alet}},
	\bibinfo{author}{\bibfnamefont{D.}~\bibnamefont{Doblar}},
	\bibinfo{author}{\bibfnamefont{A.}~\bibnamefont{Zhou}},
	\bibinfo{author}{\bibfnamefont{J.}~\bibnamefont{Tenenbaum}},
	\bibinfo{author}{\bibfnamefont{K.}~\bibnamefont{Kawaguchi}},
	\bibnamefont{and} \bibinfo{author}{\bibfnamefont{C.}~\bibnamefont{Finn}}, in
	\emph{\bibinfo{booktitle}{35th Conference on Neural Information Processing
			Systems (NeurIPS)}} (\bibinfo{year}{2021}), \eprint{arXiv:2112.03321},
	\urlprefix\url{http://arxiv.org/abs/2112.03321}.

	\bibitem[{\citenamefont{Sirignano and Spiliopoulos}(2018)}]{Sirignano2018}
	\bibinfo{author}{\bibfnamefont{J.}~\bibnamefont{Sirignano}} \bibnamefont{and}
	\bibinfo{author}{\bibfnamefont{K.}~\bibnamefont{Spiliopoulos}},
	\bibinfo{journal}{arXiv preprint}  (\bibinfo{year}{2018}),
	\eprint{arXiv:1708.07469v5}.

	\bibitem[{\citenamefont{Niarchos and Lygeros}(2006)}]{Niarchos2006}
	\bibinfo{author}{\bibfnamefont{K.~N.} \bibnamefont{Niarchos}} \bibnamefont{and}
	\bibinfo{author}{\bibfnamefont{J.}~\bibnamefont{Lygeros}},
	\bibinfo{journal}{Proceedings of the 45th IEEE Conference on Decision and
		Control} pp. \bibinfo{pages}{6313--6318} (\bibinfo{year}{2006}).

	\bibitem[{\citenamefont{Djeridane and Lygeros}(2006)}]{Djeridane2006}
	\bibinfo{author}{\bibfnamefont{B.}~\bibnamefont{Djeridane}} \bibnamefont{and}
	\bibinfo{author}{\bibfnamefont{J.}~\bibnamefont{Lygeros}},
	\bibinfo{journal}{Proceedings of the IEEE Conference on Decision and Control}
	pp. \bibinfo{pages}{3034--3039} (\bibinfo{year}{2006}).

	\bibitem[{\citenamefont{Jiang et~al.}(2017)\citenamefont{Jiang, Chou, Chen, and
					Tomlin}}]{Jiang2017}
	\bibinfo{author}{\bibfnamefont{F.}~\bibnamefont{Jiang}},
	\bibinfo{author}{\bibfnamefont{G.}~\bibnamefont{Chou}},
	\bibinfo{author}{\bibfnamefont{M.}~\bibnamefont{Chen}}, \bibnamefont{and}
	\bibinfo{author}{\bibfnamefont{C.~J.} \bibnamefont{Tomlin}},
	\bibinfo{journal}{arXiv preprint}  (\bibinfo{year}{2017}),
	\eprint{arXiv:1611.03158v2}.

	\bibitem[{\citenamefont{{Rubies Royo} and Tomlin}(2017)}]{Royo2017}
	\bibinfo{author}{\bibfnamefont{V.}~\bibnamefont{{Rubies Royo}}}
	\bibnamefont{and} \bibinfo{author}{\bibfnamefont{C.}~\bibnamefont{Tomlin}},
	\bibinfo{journal}{arXiv preprint}  (\bibinfo{year}{2017}),
	\eprint{arXiv:1611.02739v4}.

	\bibitem[{\citenamefont{Nakamura-zimmerer
					et~al.}(2021{\natexlab{a}})\citenamefont{Nakamura-zimmerer, Gong, and
					Kang}}]{Nakamura-zimmerer2020}
	\bibinfo{author}{\bibfnamefont{T.}~\bibnamefont{Nakamura-zimmerer}},
	\bibinfo{author}{\bibfnamefont{Q.~I.} \bibnamefont{Gong}}, \bibnamefont{and}
	\bibinfo{author}{\bibfnamefont{W.~E.~I.} \bibnamefont{Kang}},
	\bibinfo{journal}{arXiv preprint}  (\bibinfo{year}{2021}{\natexlab{a}}),
	\eprint{arXiv:1907.05317v5}.

	\bibitem[{\citenamefont{Nakamura-zimmerer
					et~al.}(2021{\natexlab{b}})\citenamefont{Nakamura-zimmerer, Gong, and
					Kang}}]{Nakamura-zimmerer2020a}
	\bibinfo{author}{\bibfnamefont{T.}~\bibnamefont{Nakamura-zimmerer}},
	\bibinfo{author}{\bibfnamefont{Q.}~\bibnamefont{Gong}}, \bibnamefont{and}
	\bibinfo{author}{\bibfnamefont{W.}~\bibnamefont{Kang}},
	\bibinfo{journal}{IEEE Control Systems Letters} \textbf{\bibinfo{volume}{5}},
	\bibinfo{pages}{1303} (\bibinfo{year}{2021}{\natexlab{b}}).

	\bibitem[{\citenamefont{Onken et~al.}(2021)\citenamefont{Onken, Nurbekyan, Li,
					Fung, Osher, Ruthotto, and Dec}}]{Onken2021}
	\bibinfo{author}{\bibfnamefont{D.}~\bibnamefont{Onken}},
	\bibinfo{author}{\bibfnamefont{L.}~\bibnamefont{Nurbekyan}},
	\bibinfo{author}{\bibfnamefont{X.}~\bibnamefont{Li}},
	\bibinfo{author}{\bibfnamefont{S.~W.} \bibnamefont{Fung}},
	\bibinfo{author}{\bibfnamefont{S.}~\bibnamefont{Osher}},
	\bibinfo{author}{\bibfnamefont{L.}~\bibnamefont{Ruthotto}}, \bibnamefont{and}
	\bibinfo{author}{\bibfnamefont{O.~C.} \bibnamefont{Dec}},
	\bibinfo{journal}{arXiv preprint}  (\bibinfo{year}{2021}),
	\eprint{arXiv:2104.03270v2}.

	\bibitem[{\citenamefont{Bansal and Tomlin}(2020)}]{Bansal2021}
	\bibinfo{author}{\bibfnamefont{S.}~\bibnamefont{Bansal}} \bibnamefont{and}
	\bibinfo{author}{\bibfnamefont{C.~J.} \bibnamefont{Tomlin}},
	\bibinfo{journal}{arXiv preprint}  (\bibinfo{year}{2020}),
	\eprint{arXiv:2011.02082}.

	\bibitem[{\citenamefont{Han et~al.}(2018)\citenamefont{Han, Jentzen, and
					Weinan}}]{Han2018}
	\bibinfo{author}{\bibfnamefont{J.}~\bibnamefont{Han}},
	\bibinfo{author}{\bibfnamefont{A.}~\bibnamefont{Jentzen}}, \bibnamefont{and}
	\bibinfo{author}{\bibfnamefont{E.}~\bibnamefont{Weinan}},
	\bibinfo{journal}{Proc. Natl. Acad. Sci. U. S. A.}
	\textbf{\bibinfo{volume}{115}} (\bibinfo{year}{2018}).

	\bibitem[{\citenamefont{Hure et~al.}(2020)\citenamefont{Hure, Pham, and
					Warin}}]{Hure2020}
	\bibinfo{author}{\bibfnamefont{C.}~\bibnamefont{Hure}},
	\bibinfo{author}{\bibfnamefont{H.}~\bibnamefont{Pham}}, \bibnamefont{and}
	\bibinfo{author}{\bibfnamefont{X.}~\bibnamefont{Warin}},
	\bibinfo{journal}{arXiv preprint}  (\bibinfo{year}{2020}),
	\eprint{arXiv:1902.01599v2}.

	\bibitem[{\citenamefont{Bachouch et~al.}(2022)\citenamefont{Bachouch,
	Hur{\'{e}}, Langren{\'{e}}, and Pham}}]{Bachouch2022}
	\bibinfo{author}{\bibfnamefont{A.}~\bibnamefont{Bachouch}},
	\bibinfo{author}{\bibfnamefont{C.}~\bibnamefont{Hur{\'{e}}}},
	\bibinfo{author}{\bibfnamefont{N.}~\bibnamefont{Langren{\'{e}}}},
	\bibnamefont{and} \bibinfo{author}{\bibfnamefont{H.}~\bibnamefont{Pham}},
	\bibinfo{journal}{Methodology and Computing in Applied Probability}
	\textbf{\bibinfo{volume}{24}}, \bibinfo{pages}{143} (\bibinfo{year}{2022}),
	\eprint{1812.05916}.

	\bibitem[{\citenamefont{B{\"{o}}ttcher
	et~al.}(2022)\citenamefont{B{\"{o}}ttcher, Antulov-Fantulin, and
	Asikis}}]{Bottcher2022}
	\bibinfo{author}{\bibfnamefont{L.}~\bibnamefont{B{\"{o}}ttcher}},
	\bibinfo{author}{\bibfnamefont{N.}~\bibnamefont{Antulov-Fantulin}},
	\bibnamefont{and} \bibinfo{author}{\bibfnamefont{T.}~\bibnamefont{Asikis}},
	\bibinfo{journal}{Nature Communications} \textbf{\bibinfo{volume}{13}},
	\bibinfo{pages}{1} (\bibinfo{year}{2022}).

	\bibitem[{\citenamefont{Mombaur et~al.}(2010)\citenamefont{Mombaur, Truong, and
					Laumond}}]{Mombaur2010}
	\bibinfo{author}{\bibfnamefont{K.}~\bibnamefont{Mombaur}},
	\bibinfo{author}{\bibfnamefont{A.}~\bibnamefont{Truong}}, \bibnamefont{and}
	\bibinfo{author}{\bibfnamefont{J.~P.} \bibnamefont{Laumond}},
	\bibinfo{journal}{Autonomous Robots} \textbf{\bibinfo{volume}{28}},
	\bibinfo{pages}{369} (\bibinfo{year}{2010}).

	\bibitem[{\citenamefont{Puydupin-Jamin
					et~al.}(2012)\citenamefont{Puydupin-Jamin, Johnson, and
					Bretl}}]{Puydupin-Jamin2012}
	\bibinfo{author}{\bibfnamefont{A.~S.} \bibnamefont{Puydupin-Jamin}},
	\bibinfo{author}{\bibfnamefont{M.}~\bibnamefont{Johnson}}, \bibnamefont{and}
	\bibinfo{author}{\bibfnamefont{T.}~\bibnamefont{Bretl}},
	\bibinfo{journal}{Proceedings - IEEE International Conference on Robotics and
		Automation} pp. \bibinfo{pages}{531--536} (\bibinfo{year}{2012}).

	\bibitem[{\citenamefont{Englert et~al.}(2017)\citenamefont{Englert, Vien, and
					Toussaint}}]{Englert2017}
	\bibinfo{author}{\bibfnamefont{P.}~\bibnamefont{Englert}},
	\bibinfo{author}{\bibfnamefont{N.~A.} \bibnamefont{Vien}}, \bibnamefont{and}
	\bibinfo{author}{\bibfnamefont{M.}~\bibnamefont{Toussaint}},
	\bibinfo{journal}{International Journal of Robotics Research}
	\textbf{\bibinfo{volume}{36}}, \bibinfo{pages}{1474} (\bibinfo{year}{2017}).

	\bibitem[{\citenamefont{Molloy et~al.}(2018)\citenamefont{Molloy, Ford, and
					Perez}}]{Molloy2018}
	\bibinfo{author}{\bibfnamefont{T.~L.} \bibnamefont{Molloy}},
	\bibinfo{author}{\bibfnamefont{J.~J.} \bibnamefont{Ford}}, \bibnamefont{and}
	\bibinfo{author}{\bibfnamefont{T.}~\bibnamefont{Perez}},
	\bibinfo{journal}{Automatica} \textbf{\bibinfo{volume}{87}},
	\bibinfo{pages}{442} (\bibinfo{year}{2018}),
	\urlprefix\url{https://doi.org/10.1016/j.automatica.2017.09.023}.

	\bibitem[{\citenamefont{Jin et~al.}(2021)\citenamefont{Jin, Kulic, Mou, and
					Hirche}}]{Jin2021b}
	\bibinfo{author}{\bibfnamefont{W.}~\bibnamefont{Jin}},
	\bibinfo{author}{\bibfnamefont{D.}~\bibnamefont{Kulic}},
	\bibinfo{author}{\bibfnamefont{S.}~\bibnamefont{Mou}}, \bibnamefont{and}
	\bibinfo{author}{\bibfnamefont{S.}~\bibnamefont{Hirche}},
	\bibinfo{journal}{The International Journal of Robotics Research}
	\textbf{\bibinfo{volume}{40}}, \bibinfo{pages}{848} (\bibinfo{year}{2021}),
	\eprint{arXiv:1803.07696v4}.

	\bibitem[{\citenamefont{Arora and Doshi}(2021)}]{Arora2021}
	\bibinfo{author}{\bibfnamefont{S.}~\bibnamefont{Arora}} \bibnamefont{and}
	\bibinfo{author}{\bibfnamefont{P.}~\bibnamefont{Doshi}},
	\bibinfo{journal}{Artificial Intelligence} \textbf{\bibinfo{volume}{297}},
	\bibinfo{pages}{103500} (\bibinfo{year}{2021}), \eprint{1806.06877},
	\urlprefix\url{https://doi.org/10.1016/j.artint.2021.103500}.

	\bibitem[{\citenamefont{Cao and Xie}(2022)}]{Cao2022}
	\bibinfo{author}{\bibfnamefont{K.}~\bibnamefont{Cao}} \bibnamefont{and}
	\bibinfo{author}{\bibfnamefont{L.}~\bibnamefont{Xie}}, \bibinfo{journal}{IEEE
		Transactions on Neural Networks and Learning Systems} pp.
	\bibinfo{pages}{1--8} (\bibinfo{year}{2022}).

	\bibitem[{\citenamefont{Jin et~al.}(2020{\natexlab{b}})\citenamefont{Jin, Wang,
					Yang, and Mou}}]{Jin2020a}
	\bibinfo{author}{\bibfnamefont{W.}~\bibnamefont{Jin}},
	\bibinfo{author}{\bibfnamefont{Z.}~\bibnamefont{Wang}},
	\bibinfo{author}{\bibfnamefont{Z.}~\bibnamefont{Yang}}, \bibnamefont{and}
	\bibinfo{author}{\bibfnamefont{S.}~\bibnamefont{Mou}}, in
	\emph{\bibinfo{booktitle}{Advances in Neural Information Processing
			Systems}}, edited by
	\bibinfo{editor}{\bibfnamefont{H.}~\bibnamefont{Larochelle}},
	\bibinfo{editor}{\bibfnamefont{M.}~\bibnamefont{Ranzato}},
	\bibinfo{editor}{\bibfnamefont{R.}~\bibnamefont{Hadsell}},
	\bibinfo{editor}{\bibfnamefont{M.~F.} \bibnamefont{Balcan}},
	\bibnamefont{and} \bibinfo{editor}{\bibfnamefont{H.}~\bibnamefont{Lin}}
	(\bibinfo{publisher}{Curran Associates, Inc.},
	\bibinfo{year}{2020}{\natexlab{b}}), vol.~\bibinfo{volume}{33},
	\eprint{arXiv:1912.12970}.

	\bibitem[{\citenamefont{Darbon et~al.}(2020)\citenamefont{Darbon, Langlois, and
					Meng}}]{Darbon2020}
	\bibinfo{author}{\bibfnamefont{J.}~\bibnamefont{Darbon}},
	\bibinfo{author}{\bibfnamefont{G.~P.} \bibnamefont{Langlois}},
	\bibnamefont{and} \bibinfo{author}{\bibfnamefont{T.}~\bibnamefont{Meng}},
	\bibinfo{journal}{Research in the Mathematical Sciences}
	\textbf{\bibinfo{volume}{7}}, \bibinfo{pages}{1} (\bibinfo{year}{2020}),
	\urlprefix\url{https://doi.org/10.1007/s40687-020-00215-6}.

	\bibitem[{\citenamefont{Darbon et~al.}(2021)\citenamefont{Darbon, Dower, and
					Meng}}]{Darbon2021}
	\bibinfo{author}{\bibfnamefont{J.}~\bibnamefont{Darbon}},
	\bibinfo{author}{\bibfnamefont{P.~M.} \bibnamefont{Dower}}, \bibnamefont{and}
	\bibinfo{author}{\bibfnamefont{T.}~\bibnamefont{Meng}},
	\bibinfo{journal}{arXiv preprint}  (\bibinfo{year}{2021}),
	\eprint{arXiv:2105.03336v1}.

	\bibitem[{\citenamefont{Darbon and Meng}(2021)}]{Darbon2021a}
	\bibinfo{author}{\bibfnamefont{J.}~\bibnamefont{Darbon}} \bibnamefont{and}
	\bibinfo{author}{\bibfnamefont{T.}~\bibnamefont{Meng}},
	\bibinfo{journal}{Journal of Computational Physics}
	\textbf{\bibinfo{volume}{425}}, \bibinfo{pages}{109907}
	(\bibinfo{year}{2021}),
	\urlprefix\url{https://doi.org/10.1016/j.jcp.2020.109907}.

	\bibitem[{\citenamefont{Isaacs}(1965)}]{Isaacs1965}
	\bibinfo{author}{\bibfnamefont{R.}~\bibnamefont{Isaacs}},
	\emph{\bibinfo{title}{{Differential Games: A Mathematical Theory with
					Applications to Warfare and Pursuit, Control and Optimization}}}
	(\bibinfo{publisher}{John Wiley {\&} Sons, Ltd}, \bibinfo{address}{New York},
	\bibinfo{year}{1965}).

	\bibitem[{\citenamefont{Basar and Olsder}(1999)}]{Basar1999}
	\bibinfo{author}{\bibfnamefont{T.}~\bibnamefont{Basar}} \bibnamefont{and}
	\bibinfo{author}{\bibfnamefont{G.~J.} \bibnamefont{Olsder}},
	\emph{\bibinfo{title}{{Dynamic Noncooperative Game Theory}}}
	(\bibinfo{publisher}{Society for Industrial and Applied Mathematics},
	\bibinfo{address}{Philadelphia}, \bibinfo{year}{1999}),
	\bibinfo{edition}{2nd} ed.

	\bibitem[{\citenamefont{Nash}(1951)}]{Nash1951}
	\bibinfo{author}{\bibfnamefont{J.}~\bibnamefont{Nash}}, \bibinfo{journal}{The
		Annals of Mathematics} \textbf{\bibinfo{volume}{54}}, \bibinfo{pages}{286}
	(\bibinfo{year}{1951}).

	\bibitem[{\citenamefont{Arnol'd}(1989)}]{Arnold1989}
	\bibinfo{author}{\bibfnamefont{V.~I.} \bibnamefont{Arnol'd}},
	\emph{\bibinfo{title}{{Mathematical Methods of Classical Mechanics}}}
	(\bibinfo{publisher}{Springer}, \bibinfo{address}{New York},
	\bibinfo{year}{1989}), \bibinfo{edition}{2nd} ed.

	\bibitem[{\citenamefont{Goldstein et~al.}(2001)\citenamefont{Goldstein, Poole,
					and Safko}}]{Goldstein2001}
	\bibinfo{author}{\bibfnamefont{H.}~\bibnamefont{Goldstein}},
	\bibinfo{author}{\bibfnamefont{C.}~\bibnamefont{Poole}}, \bibnamefont{and}
	\bibinfo{author}{\bibfnamefont{J.}~\bibnamefont{Safko}},
	\emph{\bibinfo{title}{{Classical Mechanics}}} (\bibinfo{publisher}{Addison
		Wesley}, \bibinfo{address}{San Franscisco}, \bibinfo{year}{2001}),
	\bibinfo{edition}{3rd} ed.

	\bibitem[{\citenamefont{Sussman and Wisdom}(2014)}]{sicm}
	\bibinfo{author}{\bibfnamefont{G.~J.} \bibnamefont{Sussman}} \bibnamefont{and}
	\bibinfo{author}{\bibfnamefont{J.}~\bibnamefont{Wisdom}},
	\emph{\bibinfo{title}{{Structure and Interpretation of Classical Mechanics}}}
	(\bibinfo{publisher}{MIT Press}, \bibinfo{address}{Cambridge},
	\bibinfo{year}{2014}), \bibinfo{edition}{2nd} ed.

	\bibitem[{\citenamefont{Lenhart and Workman}(2007)}]{Lenhart2007}
	\bibinfo{author}{\bibfnamefont{S.}~\bibnamefont{Lenhart}} \bibnamefont{and}
	\bibinfo{author}{\bibfnamefont{J.~T.} \bibnamefont{Workman}},
	\emph{\bibinfo{title}{{Optimal Control Applied to Biological Models}}}
	(\bibinfo{publisher}{CRC Press}, \bibinfo{address}{Boca Raton},
	\bibinfo{year}{2007}).

	\bibitem[{\citenamefont{Gelfand and Fomin}(2000)}]{Gelfand2000}
	\bibinfo{author}{\bibfnamefont{I.~M.} \bibnamefont{Gelfand}} \bibnamefont{and}
	\bibinfo{author}{\bibfnamefont{S.~V.} \bibnamefont{Fomin}},
	\emph{\bibinfo{title}{{Calculus of Variations}}} (\bibinfo{publisher}{Dover
		Publications, Inc.}, \bibinfo{address}{Mineola, New York},
	\bibinfo{year}{2000}).

	\bibitem[{\citenamefont{Kermack and McKendrick}(1927)}]{Petard1938}
	\bibinfo{author}{\bibfnamefont{W.~O.} \bibnamefont{Kermack}} \bibnamefont{and}
	\bibinfo{author}{\bibfnamefont{A.~G.} \bibnamefont{McKendrick}},
	\bibinfo{journal}{Proceedings of the Royal Society of London. Series A,
		Containing Papers of a Mathematical and Physical Character}
	\textbf{\bibinfo{volume}{115}}, \bibinfo{pages}{700} (\bibinfo{year}{1927}).

	\bibitem[{\citenamefont{Reluga}(2010)}]{Reluga2010}
	\bibinfo{author}{\bibfnamefont{T.~C.} \bibnamefont{Reluga}},
	\bibinfo{journal}{PLoS Computational Biology} \textbf{\bibinfo{volume}{6}},
	\bibinfo{pages}{1} (\bibinfo{year}{2010}).

	\bibitem[{\citenamefont{Schnyder et~al.}(2022)\citenamefont{Schnyder, Molina,
					Yamamoto, and Turner}}]{Schnyder2022}
	\bibinfo{author}{\bibfnamefont{S.~K.} \bibnamefont{Schnyder}},
	\bibinfo{author}{\bibfnamefont{J.~J.} \bibnamefont{Molina}},
	\bibinfo{author}{\bibfnamefont{R.}~\bibnamefont{Yamamoto}}, \bibnamefont{and}
	\bibinfo{author}{\bibfnamefont{M.~S.} \bibnamefont{Turner}},
	\bibinfo{journal}{under review}  (\bibinfo{year}{2022}).

	\bibitem[{\citenamefont{Wang et~al.}(2016)\citenamefont{Wang, Bauch,
	Bhattacharyya, D'Onofrio, Manfredi, Perc, Perra, Salath{\'{e}}, and
	Zhao}}]{Wang2016}
	\bibinfo{author}{\bibfnamefont{Z.}~\bibnamefont{Wang}},
	\bibinfo{author}{\bibfnamefont{C.~T.} \bibnamefont{Bauch}},
	\bibinfo{author}{\bibfnamefont{S.}~\bibnamefont{Bhattacharyya}},
	\bibinfo{author}{\bibfnamefont{A.}~\bibnamefont{D'Onofrio}},
	\bibinfo{author}{\bibfnamefont{P.}~\bibnamefont{Manfredi}},
	\bibinfo{author}{\bibfnamefont{M.}~\bibnamefont{Perc}},
	\bibinfo{author}{\bibfnamefont{N.}~\bibnamefont{Perra}},
	\bibinfo{author}{\bibfnamefont{M.}~\bibnamefont{Salath{\'{e}}}},
	\bibnamefont{and} \bibinfo{author}{\bibfnamefont{D.}~\bibnamefont{Zhao}},
	\bibinfo{journal}{Physics Reports} \textbf{\bibinfo{volume}{664}},
	\bibinfo{pages}{1} (\bibinfo{year}{2016}), \eprint{arXiv:1608.09010},
	\urlprefix\url{http://dx.doi.org/10.1016/j.physrep.2016.10.006}.

	\bibitem[{\citenamefont{Mcadams}(2020)}]{McAdams2020}
	\bibinfo{author}{\bibfnamefont{D.}~\bibnamefont{Mcadams}},
	\bibinfo{journal}{Covid Economics (forthcoming)}  (\bibinfo{year}{2020}),
	\urlprefix\url{https://www.ssrn.com/abstract=3593272}.

	\bibitem[{\citenamefont{Makris and Toxvaerd}(2020)}]{Makris2020}
	\bibinfo{author}{\bibfnamefont{M.}~\bibnamefont{Makris}} \bibnamefont{and}
	\bibinfo{author}{\bibfnamefont{F.}~\bibnamefont{Toxvaerd}},
	\bibinfo{journal}{Cambridge Working Papers in Economics}
	\textbf{\bibinfo{volume}{2097}} (\bibinfo{year}{2020}),
	\urlprefix\url{https://www.repository.cam.ac.uk/handle/1810/315201}.

	\bibitem[{\citenamefont{Toxvaerd}(2019)}]{Toxvaerd2019}
	\bibinfo{author}{\bibfnamefont{F.}~\bibnamefont{Toxvaerd}},
	\bibinfo{journal}{International Economic Review}
	\textbf{\bibinfo{volume}{60}}, \bibinfo{pages}{1737} (\bibinfo{year}{2019}).

	\bibitem[{\citenamefont{Rowthorn and Toxvaerd}(2020)}]{Rowthorn2020}
	\bibinfo{author}{\bibfnamefont{R.}~\bibnamefont{Rowthorn}} \bibnamefont{and}
	\bibinfo{author}{\bibfnamefont{F.}~\bibnamefont{Toxvaerd}},
	\bibinfo{journal}{Cambridge Working Papers in Economics}
	\textbf{\bibinfo{volume}{2027}} (\bibinfo{year}{2020}),
	\urlprefix\url{https://www.repository.cam.ac.uk/handle/1810/305399}.

	\bibitem[{\citenamefont{Bethune and Korinek}(2020)}]{Bethune2020}
	\bibinfo{author}{\bibfnamefont{Z.~A.} \bibnamefont{Bethune}} \bibnamefont{and}
	\bibinfo{author}{\bibfnamefont{A.}~\bibnamefont{Korinek}},
	\bibinfo{journal}{NBER Working Paper Series} \textbf{\bibinfo{volume}{27009}}
	(\bibinfo{year}{2020}), \urlprefix\url{http://www.nber.org/papers/w27009}.

	\bibitem[{\citenamefont{Eichenbaum et~al.}(2021)\citenamefont{Eichenbaum,
					Rebelo, and Trabandt}}]{Eichenbaum2021}
	\bibinfo{author}{\bibfnamefont{M.~S.} \bibnamefont{Eichenbaum}},
	\bibinfo{author}{\bibfnamefont{S.}~\bibnamefont{Rebelo}}, \bibnamefont{and}
	\bibinfo{author}{\bibfnamefont{M.}~\bibnamefont{Trabandt}},
	\bibinfo{journal}{Review of Financial Studies} \textbf{\bibinfo{volume}{34}},
	\bibinfo{pages}{5149} (\bibinfo{year}{2021}).

	\bibitem[{\citenamefont{Shaier et~al.}(2021)\citenamefont{Shaier, Raissi, and
					Seshaiyer}}]{Shaier2021}
	\bibinfo{author}{\bibfnamefont{S.}~\bibnamefont{Shaier}},
	\bibinfo{author}{\bibfnamefont{M.}~\bibnamefont{Raissi}}, \bibnamefont{and}
	\bibinfo{author}{\bibfnamefont{P.}~\bibnamefont{Seshaiyer}},
	\bibinfo{journal}{arXiv preprint}  (\bibinfo{year}{2021}),
	\eprint{arXiv:2110.05445}, \urlprefix\url{http://arxiv.org/abs/2110.05445}.

	\bibitem[{\citenamefont{Jin et~al.}(2022{\natexlab{b}})\citenamefont{Jin,
	Murphey, Kuli{\'{c}}, Ezer, and Mou}}]{Jin2020b}
	\bibinfo{author}{\bibfnamefont{W.}~\bibnamefont{Jin}},
	\bibinfo{author}{\bibfnamefont{T.~D.} \bibnamefont{Murphey}},
	\bibinfo{author}{\bibfnamefont{D.}~\bibnamefont{Kuli{\'{c}}}},
	\bibinfo{author}{\bibfnamefont{N.}~\bibnamefont{Ezer}}, \bibnamefont{and}
	\bibinfo{author}{\bibfnamefont{S.}~\bibnamefont{Mou}},
	\bibinfo{journal}{arXiv preprint}  (\bibinfo{year}{2022}{\natexlab{b}}),
	\eprint{arXiv:2008.02159v2}, \urlprefix\url{http://arxiv.org/abs/2008.02159}.

	\bibitem[{\citenamefont{Margossian and Betancourt}(2021)}]{Margossian2021}
	\bibinfo{author}{\bibfnamefont{C.}~\bibnamefont{Margossian}} \bibnamefont{and}
	\bibinfo{author}{\bibfnamefont{M.}~\bibnamefont{Betancourt}},
	\bibinfo{journal}{arXiv preprint}  (\bibinfo{year}{2021}),
	\eprint{arXiv:2112.14217}, \urlprefix\url{http://arxiv.org/abs/2112.14217}.

	\bibitem[{\citenamefont{Blondel et~al.}(2021)\citenamefont{Blondel, Berthet,
	Cuturi, Frostig, Hoyer, Llinares-L{\'{o}}pez, Pedregosa, and
	Vert}}]{Blondel2021}
	\bibinfo{author}{\bibfnamefont{M.}~\bibnamefont{Blondel}},
	\bibinfo{author}{\bibfnamefont{Q.}~\bibnamefont{Berthet}},
	\bibinfo{author}{\bibfnamefont{M.}~\bibnamefont{Cuturi}},
	\bibinfo{author}{\bibfnamefont{R.}~\bibnamefont{Frostig}},
	\bibinfo{author}{\bibfnamefont{S.}~\bibnamefont{Hoyer}},
	\bibinfo{author}{\bibfnamefont{F.}~\bibnamefont{Llinares-L{\'{o}}pez}},
	\bibinfo{author}{\bibfnamefont{F.}~\bibnamefont{Pedregosa}},
	\bibnamefont{and} \bibinfo{author}{\bibfnamefont{J.-P.} \bibnamefont{Vert}},
	\bibinfo{journal}{arXiv preprint}  (\bibinfo{year}{2021}),
	\eprint{arXiv:2105.15183}, \urlprefix\url{http://arxiv.org/abs/2105.15183}.

	\bibitem[{Note1()}]{Note1}
	Note1, \bibinfo{note}{at the time of writing, the JAXopt\cite {Blondel2021}
		library only supported root-finding using a simple bisection algorithm, which
		required specifying a bracketing interval around the root. This made it
		impractical to use during training.}

	\bibitem[{\citenamefont{Kingma and Ba}(2014)}]{Kingma2015}
	\bibinfo{author}{\bibfnamefont{D.~P.} \bibnamefont{Kingma}} \bibnamefont{and}
	\bibinfo{author}{\bibfnamefont{J.~L.} \bibnamefont{Ba}}, in
	\emph{\bibinfo{booktitle}{3rd International Conference on Learning
			Representations (ICLR 2015)}} (\bibinfo{year}{2014}), pp.
	\bibinfo{pages}{1--15}, \urlprefix\url{https://arxiv.org/abs/1412.6980}.

	\bibitem[{\citenamefont{Bradbury et~al.}(2018)\citenamefont{Bradbury, Frostig,
					Hawkins, Johnson, Leary, Maclaurin, Necula, Paszke, VanderPlas,
					Wanderman-Milne et~al.}}]{Bradbury2018}
	\bibinfo{author}{\bibfnamefont{J.}~\bibnamefont{Bradbury}},
	\bibinfo{author}{\bibfnamefont{R.}~\bibnamefont{Frostig}},
	\bibinfo{author}{\bibfnamefont{P.}~\bibnamefont{Hawkins}},
	\bibinfo{author}{\bibfnamefont{M.~J.} \bibnamefont{Johnson}},
	\bibinfo{author}{\bibfnamefont{C.}~\bibnamefont{Leary}},
	\bibinfo{author}{\bibfnamefont{D.}~\bibnamefont{Maclaurin}},
	\bibinfo{author}{\bibfnamefont{G.}~\bibnamefont{Necula}},
	\bibinfo{author}{\bibfnamefont{A.}~\bibnamefont{Paszke}},
	\bibinfo{author}{\bibfnamefont{J.}~\bibnamefont{VanderPlas}},
	\bibinfo{author}{\bibfnamefont{S.}~\bibnamefont{Wanderman-Milne}},
	\bibnamefont{et~al.}, \emph{\bibinfo{title}{{JAX: composable transformations
					of Python+NumPy Programs}}} (\bibinfo{year}{2018}).

	\bibitem[{\citenamefont{Hunter}(2007)}]{Hunter2007}
	\bibinfo{author}{\bibfnamefont{J.~D.} \bibnamefont{Hunter}},
	\bibinfo{journal}{Computing in Science and Engineering}
	\textbf{\bibinfo{volume}{9}}, \bibinfo{pages}{90} (\bibinfo{year}{2007}).

\end{thebibliography}

\end{document}